\documentclass[10pt,twocolumn,letterpaper]{article}

\usepackage{cvpr}              

\usepackage{booktabs}
\usepackage[table,xcdraw]{xcolor}
\usepackage{pifont}
\usepackage{times}
\usepackage{epsfig}
\usepackage{graphicx}
\usepackage{amsmath}
\usepackage{amssymb}
\usepackage{multirow}
\usepackage{tabularx}
\usepackage{dsfont}
\usepackage{url}
\usepackage{float}
\usepackage{color}
\usepackage{xcolor}
\usepackage{colortbl}
\usepackage{amsthm}
\usepackage[linesnumbered,titlenumbered,ruled,vlined,commentsnumbered,noend]{algorithm2e}
\usepackage[export]{adjustbox}
\usepackage{comment}
\usepackage{flushend}
\usepackage{indentfirst}
\usepackage[accsupp]{axessibility}
\usepackage[utf8]{inputenc} 
\usepackage[T1]{fontenc}    
\usepackage{url}            
\usepackage{enumitem}
\usepackage{booktabs}       
\usepackage{amsfonts}       
\usepackage{nicefrac}       
\usepackage{microtype}      
\usepackage[rightcaption]{sidecap}
\usepackage{wrapfig}

\usepackage[pagebackref,breaklinks,citecolor=citecolor,colorlinks]{hyperref}

\definecolor{dg}{rgb}{0,0.694,0.298}
\definecolor{purple}{rgb}{0.4,0.176,0.569}
\definecolor{Gray}{gray}{0.6}
\definecolor{citecolor}{RGB}{65,105,225}

\usepackage[capitalize]{cleveref}
\crefname{section}{Sec.}{Secs.}
\Crefname{section}{Section}{Sections}
\Crefname{table}{Table}{Tables}
\crefname{table}{Tab.}{Tabs.}

\usepackage{xspace}
\makeatletter
\DeclareRobustCommand\onedot{\futurelet\@let@token\@onedot}
\def\@onedot{\ifx\@let@token.\else.\null\fi\xspace}
 
\def\ie{\emph{i.e}\onedot}

\def\etal{\emph{et al}\onedot}
\makeatother

\AtBeginDocument{%
  \providecommand\BibTeX{{%
    \normalfont B\kern-0.5em{\scshape i\kern-0.25em b}\kern-0.8em\TeX}}}

\def\cvprPaperID{3104} 
\def\confName{CVPR}
\def\confYear{2023}

\begin{document}


\title{ Evading DeepFake Detectors via Adversarial Statistical Consistency}

\author{Yang Hou\textsuperscript{1},\quad Qing Guo\textsuperscript{2,3}\thanks{Corresponding author: tsingqguo@ieee.org}, \quad Yihao Huang\textsuperscript{4}, \quad Xiaofei Xie \textsuperscript{5}, \quad Lei Ma\textsuperscript{6,7}, \quad Jianjun Zhao\textsuperscript{1}\\
~\\
\textsuperscript{1}Kyushu University, Japan, \quad
\textsuperscript{2}Centre for Frontier AI Research (CFAR), A*STAR, Singapore,\\
\textsuperscript{3}Institute of High Performance Computing (IHPC), A*STAR, Singapore,\\
\textsuperscript{4}Nanyang Technological University, Singapore, \quad
\textsuperscript{5}Singapore Management University, Singapore\\
\textsuperscript{6}University of Alberta, Canada, \quad
\textsuperscript{7}The University of Tokyo, Japan
}
\maketitle
\thispagestyle{empty}


\begin{abstract}
   In recent years, as various realistic face forgery techniques known as DeepFake improves by leaps and bounds, more and more DeepFake detection techniques have been proposed. These methods typically rely on detecting statistical differences between natural (i.e., real) and DeepFake-generated images in both spatial and frequency domains.
   In this work, we propose to explicitly minimize the statistical differences  to evade state-of-the-art DeepFake detectors. 
   To this end, we propose a statistical consistency attack (StatAttack) against DeepFake detectors, which contains two main parts.
   First, we select several statistical-sensitive natural degradations (i.e., exposure, blur, and noise) and add them to the fake images in an adversarial way.
   Second, we find that the statistical differences between natural and DeepFake images are positively associated with the distribution shifting between the two kinds of images, and we propose to use a distribution-aware loss to guide the optimization of different degradations. As a result, the feature distributions of generated adversarial examples is close to the natural images.
   Furthermore, we extend the StatAttack to a more powerful version, MStatAttack, where we extend the single-layer degradation to multi-layer degradations sequentially and use the loss to tune the combination weights jointly.
   Comprehensive experimental results on four spatial-based detectors and two frequency-based detectors with four datasets demonstrate the effectiveness of our proposed attack method in both white-box and black-box settings.
   %
\end{abstract}

\graphicspath{{./}}
\begin{figure}[t]
  \centering
  \includegraphics[width=\linewidth]{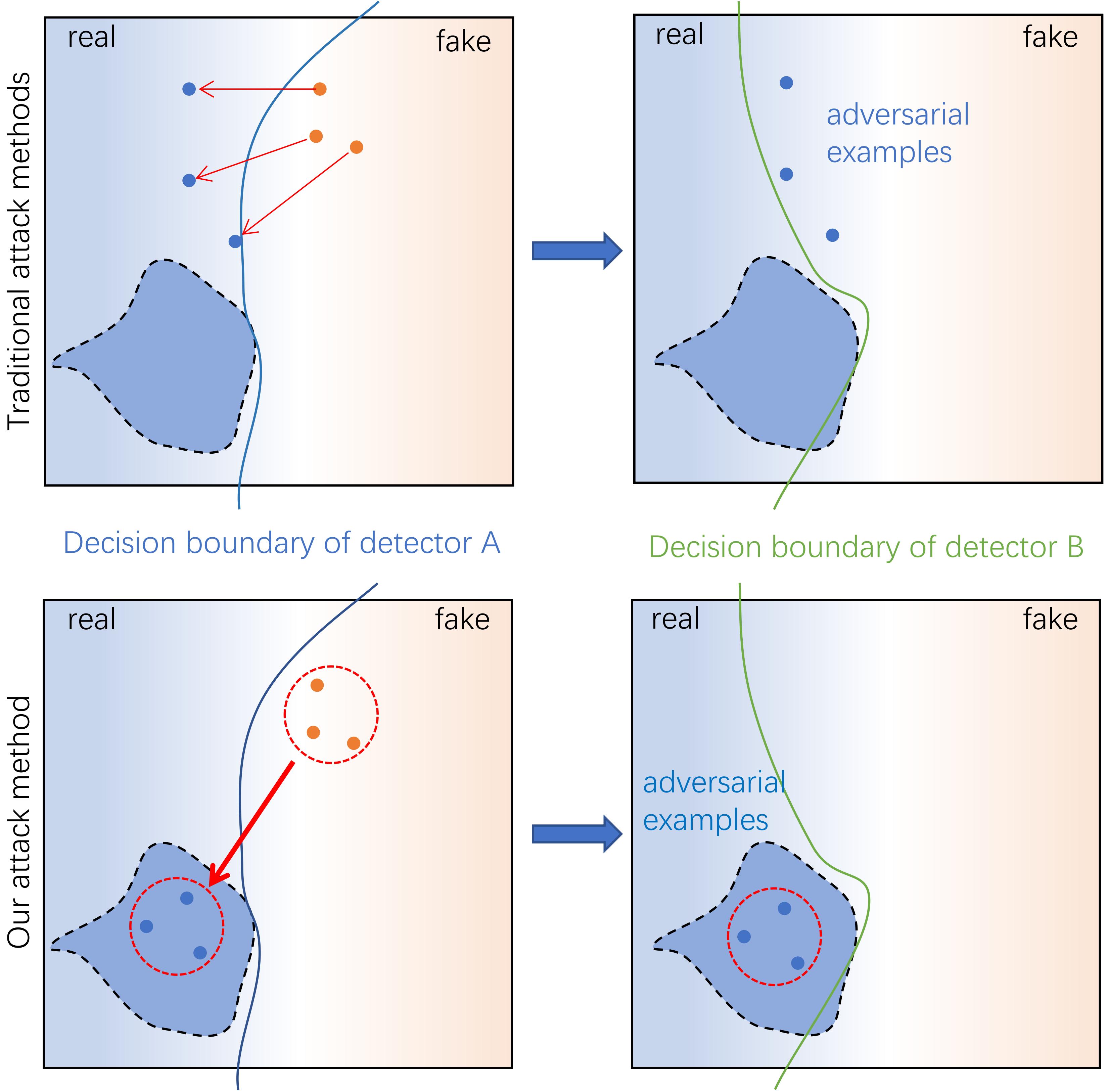}
   \caption{ 
   Principle of our method. 
   The light blue region and the light red region represent the embedding space of natural/real images and fake images, respectively. 
   The dark blue region represents the embedding spaces of real images shared by different detectors. 
   The first row shows that a typical attack can map the fake samples (\ie, the orange points) to the `real' samples that can fool detector A but fail to mislead detector B. 
   The second row shows that our method is to map the fake samples to the common regions of different detectors, which can fool both detectors.
   }
   \label{fig:fig1}
\end{figure}
\section{Introduction}
\label{sec:intro}
Recent advances in facial generation and manipulation using deep generative approaches (\ie, DeepFake \cite{juefei2022countering}) have attracted considerable media and public attentions. Fake images can be easily created using a variety of free and open-source tools. However, the misuse of DeepFake raises security and privacy concerns, particularly in areas such as politics and pornography \cite{DeepFake,thies2019deferred}. The majority of back-end technologies for DeepFake rely on generative adversarial networks (GANs). As GANs continue to advance, state-of-the-art (SOTA) DeepFake has achieved a level of sophistication that is virtually indistinguishable to human eyes.

Although these realistic fake images can spoof human eyes, SOTA DeepFake detectors can still effectively detect subtle `fake features' by leveraging the powerful feature extraction capabilities of deep neural networks (DNNs). 
However, recent studies \cite{gandhi2020adversarial,carlini2020evading} have shown that these detectors are vulnerable to adversarial attacks that can bypass detectors by injecting perturbations into fake images. Additionally, adversarial examples pose a practical threat to DeepFake detection if they can be transferred between different detectors. Adversarial attacks are often used to verify the robustness of DeepFake detectors \cite{hussain2021adversarial}. Therefore, in order to further research and develop robust DeepFake detectors, it is crucial to develop effective and transferable adversarial attacks.

Several previous studies have explored the transferability of adversarial examples \cite{xie2019improving,wang2021enhancing}, which refers to the ability of adversarial examples designed for a specific victim model to attack other models trained for the same task. However, achieving transferable attacks against DeepFake detectors is particularly challenging due to variations in network architectures and training examples caused by different data augmentation and pre-processing methods. 
These differences often result in poor transferability of adversarial examples crafted from typical attack methods when faced with different DeepFake detectors.

Current detection methods typically rely on detecting statistical differences in spatial and frequency domains between natural and DeepFake-generated images, (as explained in Section \ref{sec:sec3}), and various detectors share some common statistical properties of natural images \cite{mccloskey2019detecting, durall2020watch, zhang2019detecting}. 
These prior knowledge and discoveries inspire us to design an attack method with strong transferability that can minimize statistical differences explicitly. 
Toward the transferable attack, we propose a novel attack method, StatAttack. Specifically, we select three types of statistical-sensitive natural degradations, including exposure, blur, and noise, and add them to fake images in an adversarial manner.
In addition, our analysis indicates that the statistical differences between the real and fake image sets are positively associated with their distribution shifting.
Hence, we propose to mitigate these differences by minimizing the distance between the feature distributions of fake images and that of natural images. To achieve this, we introduce a novel distribution-aware loss function that effectively minimizes the statistical differences.
Figure \ref{fig:fig1} illustrates the principle of our proposed attack method.
Moreover, We expand our attack method to a more powerful version, MStatAttack. This improved approach performs multi-layer degradations and can dynamically adjust the weights of each degradation in different layers during each attack step. With the MStatAttack, we can develop more effective attack strategies and generate adversarial examples that appear more natural.

Our contributions can be summarized as the following: 
\begin{itemize}[itemsep=2pt,topsep=0pt,parsep=0pt]
\item We propose a novel natural degradation-based attack method, StatAttack.  StatAttack can fill the feature distributions difference between real and fake images by minimizing a distribution-aware loss.

\item To enhance the StatAttack, we further propose a multi-layer counterpart, MStatAttack, which can select a more effective combination of perturbations and generate more natural-looking adversarial examples.

\item We conduct comprehensive experiments on four spatial-based and two frequency-based DeepFake detectors using four datasets. The experimental results demonstrate the effectiveness of our attack in both white-box and black-box settings.

\end{itemize}

\section{Related Work}

{\bf DeepFake generation.}
Generative adversarial networks (GANs) and their variants have achieved impressive results in image generation and manipulation, leading to the development of DeepFake technology. DeepFake utilizes GANs to generate various types of fake images or videos. 
The current DeepFake techniques can be roughly categorized into three categories: entire face synthesis \cite{perarnau2016invertible,karras2017progressive,karras2019style}, face identity swap \cite{gao2021information,nirkin2019fsgan}, and face manipulation \cite{liu2019stgan,he2019attgan,gao2021high}.

Entire face synthesis aims to generate realistic human faces that do not really exist, such as ProGAN \cite{karras2017progressive} and StyleGAN \cite{karras2019style} developed by NVIDIA. The face identity swap replaces a person's face with the face of the target person. FaceApp and FaceSwap are two Popular DeepFake generation tools that employ GANs to achieve identity swapping. In face manipulation, StarGAN \cite{choi2018stargan}, STGAN \cite{liu2019stgan}, A3GAN \cite{zhai2022a3gan}, and AttGAN \cite{he2019attgan} are employed to edit and manipulate the attributes and expressions of human faces, such as changing the hair color, adding beards, wearing glasses or creating smiling faces.

Malicious applications that utilize these generation methods may pose a significant threat to public information security. In order to fully evaluate the effectiveness of our attack, we conduct a comprehensive evaluation based on the generation methods described above. The generated face dataset includes entire face synthesis images, face identity swap images, and face manipulation images, allowing for a comprehensive analysis of the attack's performance.

{\bf DeepFake detection.} To prevent the misuse of DeepFake technologies, several DNN-based DeepFake detection methods have been proposed. DeepFake detection is essentially a binary classification problem that involves distinguishing between fake images and natural images. Among the existing detection methods, some works are focused on extracting the spatial domain information \cite{huang2022fakelocator, wang2020cnn,jeon2020fdftnet,qi2020deeprhythm} of images for DeepFake detection, while others explore the difference in frequency domain information \cite{frank2020leveraging,qian2020thinking,barni2020cnn} between fake images and natural images. These detection methods are entirely data-driven and aim to enable the model to learn the statistical differences between fake and natural images. With the powerful feature extraction capability of DNNs, these detection methods achieve superior detection results under their respective experimental settings. To fully evaluate the effectiveness of our attack method, we perform attack experiments on both spatial and frequency-based detectors to evaluate the effectiveness of our attacks.

\begin{figure}[t]
  \centering
  \includegraphics[width=\linewidth]{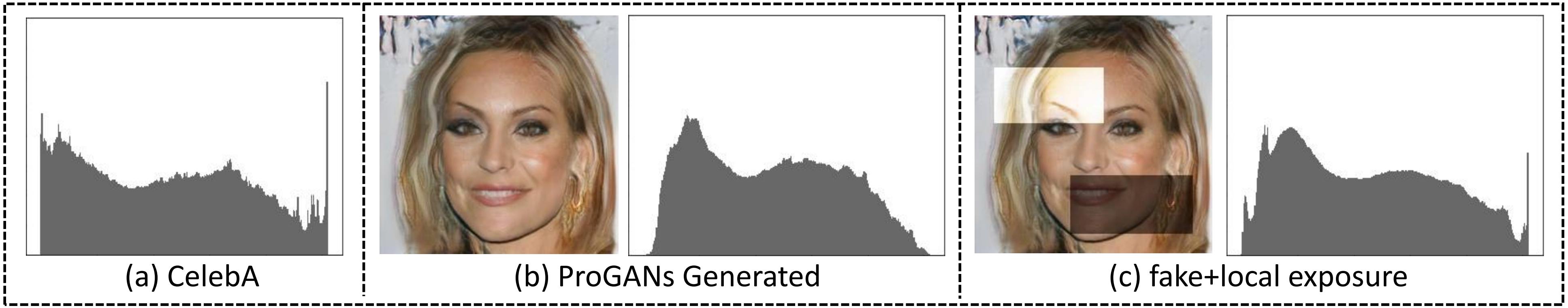}

   \caption{Brightness statistic differences. (a) shows the average brightness histogram of a natural face dataset, while (b) shows the average brightness histogram of a ProGAN-generated fake face dataset that lacks saturated and under-exposed region. (c) is the same datasets as (b) with partial exposure adjustment. After adjusting the local brightness, saturated and under-exposed pixel values in fake images histogram appear. }
   \label{fig:brightness}
\end{figure}

\begin{figure}[t]
  \centering
  \includegraphics[width=\linewidth]{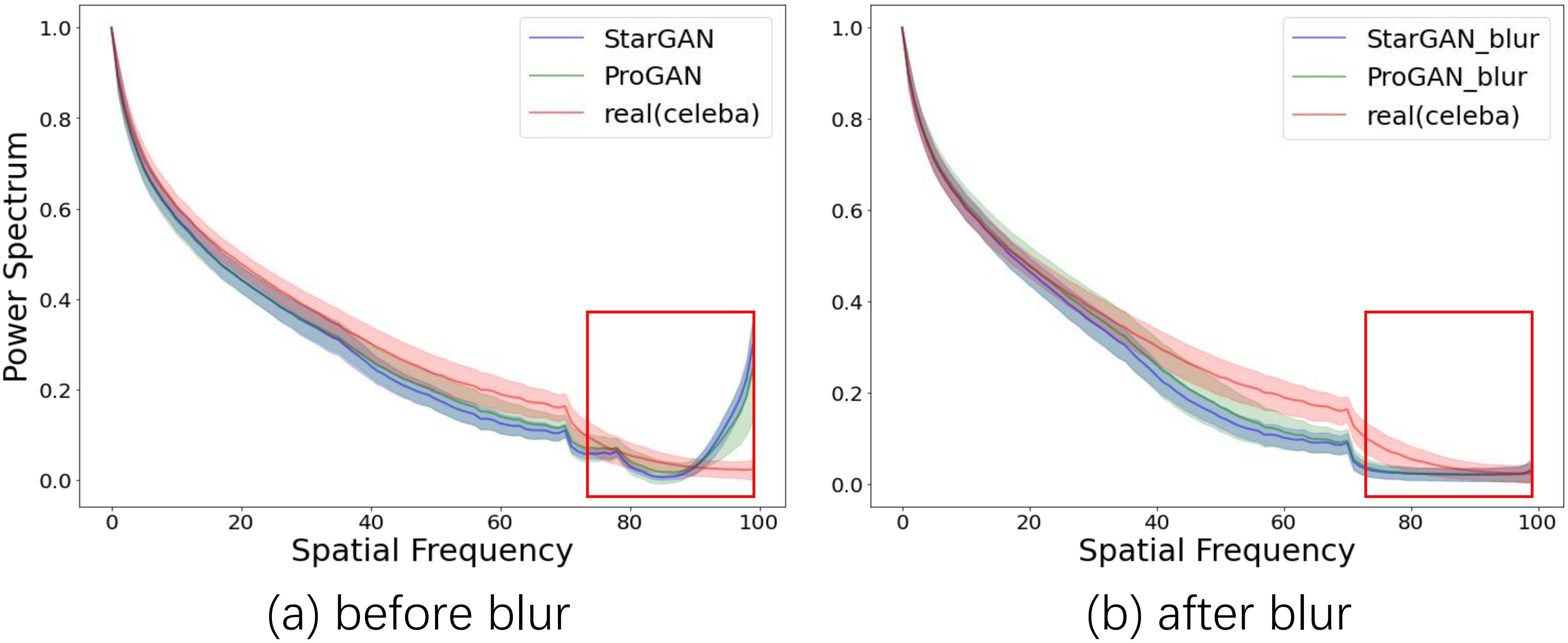}

   \caption{ Frequency Statistical differences (power spectrum). (a) shows that GAN-generated images suffer from the dramatic increment of the high-frequency components compared with natural images. (b) reveals that the Gaussian blurring process can reduce these frequency differences. }
   \label{fig:power}
\end{figure}

{\bf Adversarial attack.}
Adversarial attack aims to generate adversarial perturbations and add them to inputs. The generated adversarial examples can fool the target model into predicting an incorrect label. Some works investigate the effect of additive perturbations on classifier robustness in a white-box setting \cite{goodfellow2015explaining,madry2017towards,carlini2017towards}, while others focus on studying the transferability of the adversarial example on the same task in a black-box setting \cite{ilyas2018black,cheng2019improving,guo2019simple,shi2019curls}. More recently, some other works have tried to employ natural degradation as attack perturbations, such  as motion blur, vignetting, rain streaks, exposure, and watermark \cite{gao2022can,guo2020watch,ijcai2021p145,jia2020adv}.

For adversarial attacks against DeepFake detection, several existing works have studied the robustness of DeepFake detectors with additive perturbation-based attacks in different experimental settings \cite{neekhara2021adversarial,gandhi2020adversarial,li2021exploring,hussain2021adversarial,jia2022exploring}. Meanwhile, some works point out that adversarial examples with strong transferability can pose a practical threat to DeepFake detection \cite{neekhara2021adversarial,carlini2020evading}.  To achieve transferable attacks, we investigate the potential of natural degradation-based attacks to evade various DeepFake detectors.

\section{Statistical Differences \label{sec:sec3}}

\begin{figure}[t]
  \centering
  \includegraphics[width=\linewidth]{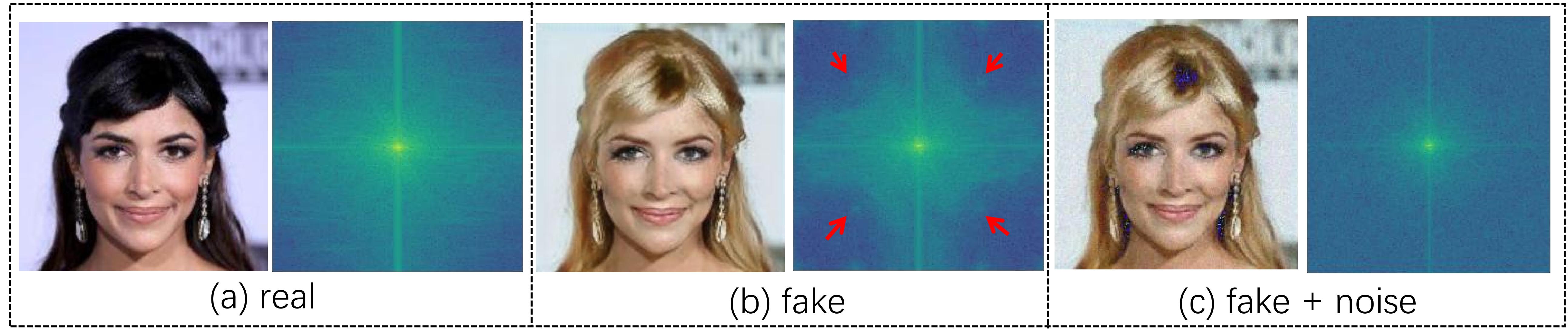}
   \caption{ Frequency Statistical differences (spectrogram). (a) is a real face image, (b) is a stargan-tampered image (Black hair turns yellow), and (c) is a fake image with random noise added. Their histograms are shown on the right of each image (Zoom in to see details). Compared to the natural image, regular frequency artifacts appear in the spectrogram of the stargan-tampered image (indicated by a red arrow). After adding noise, the high-frequency artifacts in the fake image disappear.}
   \label{fig:frequency}
\end{figure}

Several detection methods reveal statistical differences between natural and fake images in both the frequency and spatial domains. As seen in Figure \ref{fig:brightness}, McCloskey \etal \cite{mccloskey2019detecting} demonstrate that the network architecture and training process of GANs can cause differences in the brightness statistics between GAN-generated images and natural images. For example, some GANs are only capable of generating images with limited intensity values and fail to produce saturated and under-exposed regions. In addition, as illustrated in Figure \ref{fig:power}, Durall \etal \cite{durall2020watch} explore the statistical differences in frequency information between natural and GAN-generated images, revealing significant differences in high-frequency components. Similarly, as we can see from Figure \ref{fig:frequency} (a), Zhang \etal \cite{zhang2019detecting} find some statistical clues by directly observing the spectrograms of GANs-generated images. Their experimental results show that the GANs would inevitably leave regular high-frequency artifacts in the manipulated images. To mitigate these differences, we aim to add corresponding adversarial degradations.

\begin{figure*}[ht]
    \centering
    \includegraphics[width=\textwidth]{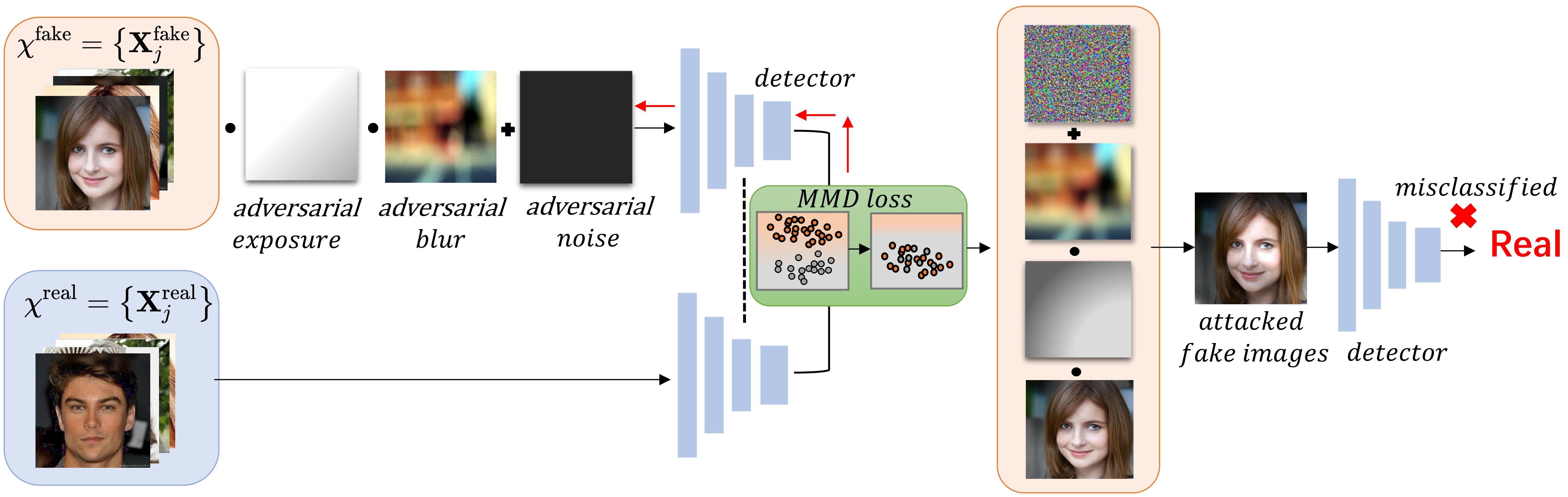}
    \caption{Pipeline of our attack. First, we add initialized adversarial degradations (adversarial exposure, adversarial blur, and adversarial noise) to the fake images.  Then, We collect the feature distributions of both natural and fake images and calculate their MMD values. Finally, we optimize the perturbation parameters by minimizing the loss function and applying the optimized degradations to fake images.}
    \label{fig:methods}
\end{figure*}
\section{Statistical Consistency Attack (StatAttack)}
In this section, we first introduce three natural degradations that reduce the statistical differences between natural and fake images, which are referred to as statistical-sensitive degradations. We then demonstrate how to adversarially add these natural degradations to the fake images. Finally, we present our objective function and explain how it reduces statistical differences effectively.
\subsection{Statistical-Sensitive Degradation} 

{\bf Selection of degradations.}
We discuss three common natural degradations that can effectively reduce the statistical differences in Section \ref{sec:sec3}:
\ding{182} Changing exposure. Adjusting the brightness values of pixels in local areas can change the brightness distribution of fake images. As shown in Figure \ref{fig:brightness} (c), oversaturating and underexposing the pixels in local areas of fake images can make their brightness distributions similar to those of natural images.
\ding{183} Adding blur. Filtering out the high-frequency components in fake images with a Gaussian low-pass filter can reduce the statistical differences in the frequency domain. As illustrated in Figure \ref{fig:power}, after performing Gaussian filtering on fake images, the distribution of high-frequency components becomes consistent with that of natural images.
\ding{184} Adding noise. Figure \ref{fig:frequency} (c) shows the spectrum of fake images after adding random noise, and we see that the regular artifacts in fake images are eliminated.
It is worth noting that adding local exposure to the image can change the information not only in the spatial domain but also in the frequency domain. Furthermore, Gaussian filtering can also reduce some regular high-frequency artifacts in the fake image.
In the following, we introduce how to add the three degradations in an adversarial way.

\vspace{0.2em}
{\bf Naturalness adversarial exposure.}
We aim to inject local exposure into the fake images, making their brightness statistics more similar to natural images. Besides, since natural-world exposure changes smoothly, the exposure injected into the fake image should also have similar properties ( \ie, adjacent pixels in an image have similar exposure values). In order to achieve both of these objectives, we adopt the adversarial exposure generation model proposed by Gao \etal. \cite{gao2022can}. This model comprises two main components: a polynomial model for generating exposure and a smoothing formula for maintaining the naturalness of the exposure. Let $\tilde{\mathbf{E}}$ represent the polynomial model for generating exposure, and $\mathbf{X}^\text{fake}$ represent a fake image. We can inject adversarial exposure into the fake image by
\begin{equation}
P_{e}(\mathbf{X}^\text{fake})=\text{log}^{-1}(\tilde{\mathbf{X}}^\text{fake}+\tilde{\mathbf{E}}),
\label{con:exposure}
\end{equation}
where the operation `$\tilde{\cdot}$' represents the logarithmic operation. The polynomial model comprises two sets of parameters, denoted by $\textbf{a}$ and $\varphi$, which are optimized to generate the adversarial exposure. To maintain the naturalness of the exposure, we constrain the values of $\textbf{a}$ and $\varphi$ by adding the smoothing equation to the loss function
\begin{equation}
    S\left ( \textbf{a},\varphi\right ) = -\lambda_{a}\left \| \textbf{a}\right \|_{2}^{2}-\lambda_{\varphi} \left \| \bigtriangledown \varphi  \right \|_{2}^{2},  
\end{equation}
where the hyper-parameters $\lambda_{a}$ and $\lambda_{\varphi}$ are used to regulate the balance between adversarial attack and smoothness.

\vspace{0.2em}
\textbf{\bf Adversarial Gaussian blur.} 
For a normal Gaussian blur, Gaussian kernels used at different locations in an image have the same value. To add Gaussian blur degradation in an adversarial way, we propose to design Gaussian kernels as learnable adversarial kernels. This approach allows the adversarial perturbations to be applied adaptively at different locations of the image \cite{guo2020watch}. 
Specifically, we denote $\sigma_{x_{i},y_{i}} $ as the initial standard deviation of the Gaussian kernel at each pixel position, where $i$ denotes the $i$-th pixel and $x_{i},y_{i}$ presents the corresponding image coordinates. We aim to learn a standard deviation map $\sigma$, where the $\sigma_{x_{i},y_{i}}$ represents the standard deviation at each pixel position. Note that, the larger $\sigma$ should result in more blurry images. With $\sigma_{x_{i},y_{i}} $, we calculate the adversarial Gaussian kernel for the $i$th pixel (\ie, $\mathbf{H}_{i}$) as follows:
\begin{equation}
\mathbf{H}_{i}(u,v) = \frac{1}{2\pi \left ( \sigma_{x_{i},y_{i}} \right )^{2}} exp\left ( -\frac{u^{2}+v^{2}}{2\pi \left ( \sigma_{x_{i},y_{i}} \right )^{2}}  \right ), 
\end{equation}
where $u$ and $v$ represent the relative coordinate of the point within the Gaussian kernel to the central pixel $(x_{i},y_{i})$.With the adversarial Gaussian kernel, we blur the fake image by
\begin{equation}
    P_{b}(\mathbf{X}^\text{fake})=\sum_{i\in \mathcal{N}(i)}^{}g(\mathbf{X}_{i}^\text{fake},k)\ast \mathbf{H}_{i}.
    \label{con:blur}
\end{equation}
In this equation, $\mathcal{N}(i)$ refers to all pixels in $\mathbf{X}_{fake}$, $g(\mathbf{X}_{i}^\text{fake},k)$ represents the region centered at the $i$th pixel and enclosed within a Gaussian kernel of radius $k$, and the term `$\ast$' denotes the filtering operation.
We can feed the generated image $P_{b}(\mathbf{X}^\text{fake})$ an adversarial-related loss function and minimize it to get $\{\sigma_{x_{i},y_{i}}\}$ and $\mathbf{H}_{i}$ via Eq. \eqref{con:blur}. 
Subsequently, we apply pixel-level Gaussian blur on the fake images through the computed Gaussian kernel.

\vspace{0.2em}
{\bf Adversarial noise.} To generate adversarial noise, we adopt a simple and effective method. The adversarial noise is denoted as follows:
\begin{equation}
    P_{n}(\mathbf{X}^\text{fake}) = \mathbf{X}^\text{fake}+\mathbf{N}_{a},
    \label{con:noise}
\end{equation}
where $\mathbf{N}_{a}$ refers to an adversarial noise map with the same size as $\mathbf{X}^\text{fake}$. During each attack step, we generate the adversarial noise by minimizing our adversarial loss function and subsequently adding it to each fake image. To ensure that the adversarial noise does not significantly impact the quality of the image, we enforce sparsity on $\mathbf{N}_{a}$ using a constraint term. This constraint term is incorporated into the final objective function.

In StatAttack, we apply these three perturbations sequentially to fake images. Based on Eq. \eqref{con:exposure}, \eqref{con:blur}, and \eqref{con:noise}, we can summarize the aforementioned perturbations as follows:
\begin{equation}
    P_{\theta }(\mathbf{X}^\text{fake}) = P_{n}(P_{b}(P_{e}(\mathbf{X}^\text{fake}))),
\end{equation}
where the $P_{\theta }(\mathbf{X}^\text{fake})$ contains four sets of parameters to be optimized, i.e., $\left \{ \textbf{a},\varphi,\sigma,\mathbf{N}_{a} \right \}$.
\begin{figure}[t]
    \centering
    \includegraphics[width=\linewidth]{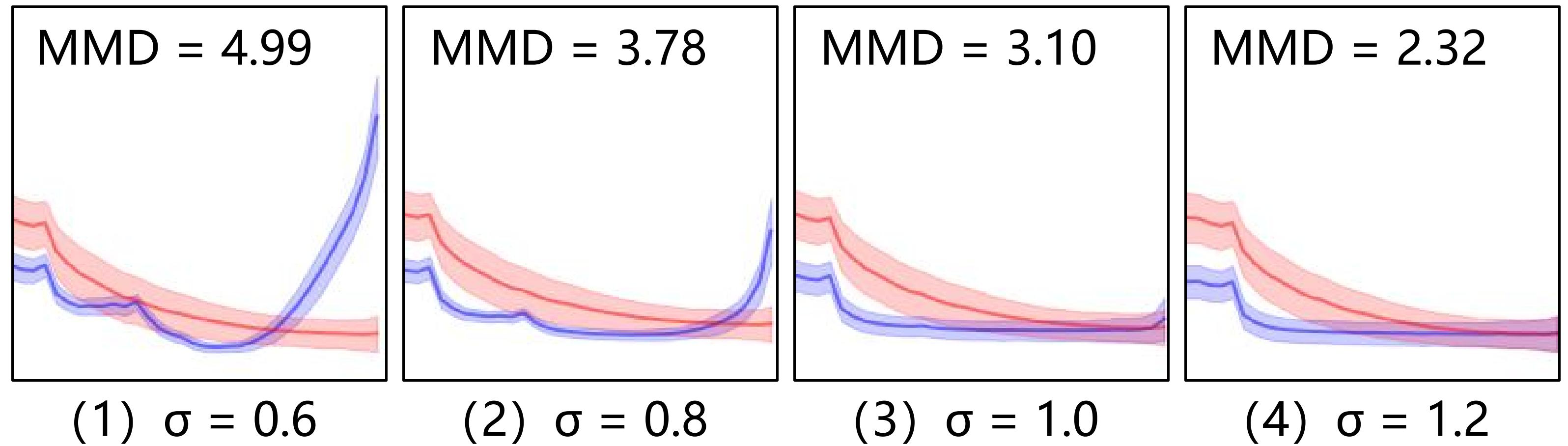}
    \caption{Correlations between statistical differences and the distribution shifting of the natural and fake image sets. (1) - (4) display the power spectrum and the MMD values between feature distributions of two sets. The blue curves represent the high-frequency components of GAN-generated images while the red curves depict natural ones. We process the fake images using Gaussian blur with various Gaussian kernels $\sigma$ and observe that as the high-frequency statistical differences decrease, the MMD values also decrease.}
    \label{fig:MMD}
\end{figure}
\subsection{Distribution-aware Statistical Consistency}
Given a set of real images and a set of fake images, we study correlations between statistical differences and distribution shifting of the two sets and find that statistical differences are positively associated with the distribution shifting between the two sets.
Figure \ref{fig:MMD} illustrates a simple experiment that we investigate this relationship.
In this experiment, we first collect a set of natural images and a set of StarGan-generated fake images, and use a Gaussian filter to handle fake images with varying $\sigma$. Then, we use maximum mean discrepancy (MMD) \cite{gretton2006kernel} to measure the distance between the feature distributions of the filtered fake images and natural images. It is worth noting that increasing the $\sigma$ can reduce the high-frequency spectrum in the fake images.
We find that the MMD value decreases as the statistical difference reduces. Moreover, previous studies \cite{wang2018improving, dziugaite2015training} have pointed out that the MMD between two feature distributions at the $i$-th layer is end-to-end differentiable, enabling the use of MMD as a loss function.

\begin{figure}[t]
    \centering
    \includegraphics[width=\linewidth]{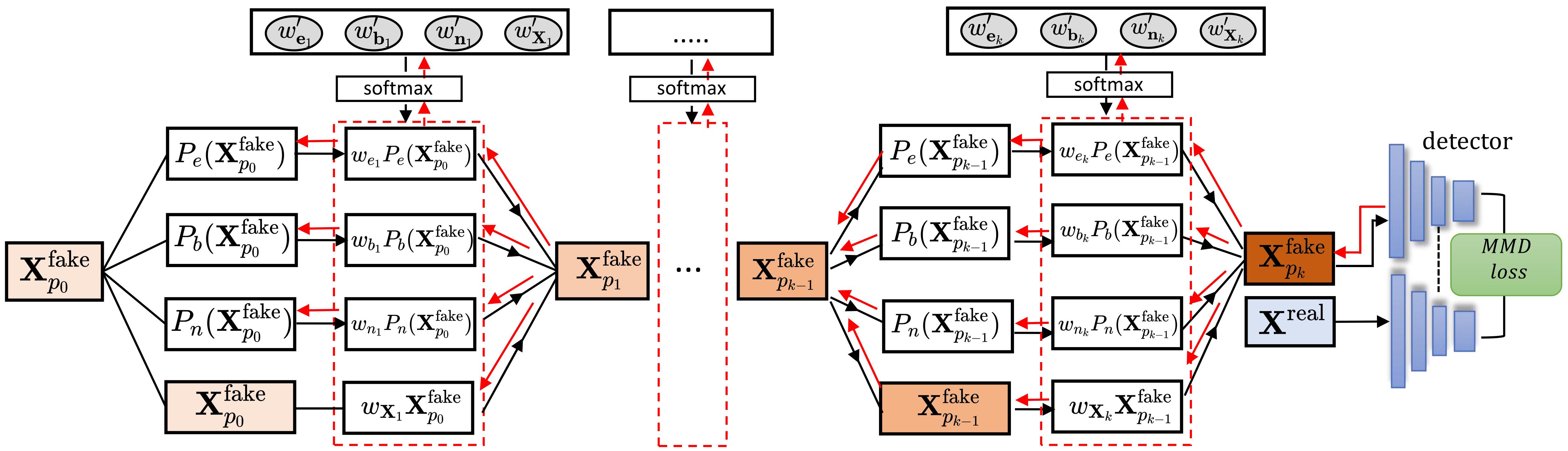}
    \caption{Architecture of MStatAttack. The weights and parameters of each attack pattern are optimized jointly when minimizing the objective function. }
    \label{fig:MStatAttack}
\end{figure}
\vspace{0.2em}
\textbf{\bf Objective function.} 
An attack is considered successful for a binary classifier if the generated adversarial example is classified as the opposite class. Our objective is to generate adversarial examples that closely resemble natural images in terms of feature distributions. Let $\chi ^\text{fake}=\left \{ \mathbf{X}^\text{fake}_{j} \right \}$ and $\chi ^\text{real}=\left \{ \mathbf{X}^\text{real}_{j} \right \}$ represent a set of fake and natural images. We define our objective function as follows:
\begin{equation}
\begin{split}
    &\underset{\textbf{a},\varphi,\sigma,\mathbf{N}_{a}}{arg\min} J_{MMD}\left ( P_{\theta}\left (  \chi ^\text{fake}\right ), \chi ^\text{real}  \right ) + S( \textbf{a},\varphi ), \\& \qquad \qquad \qquad \mathrm {subject\; to }\; \left \| \mathbf{N}_{a}   \right \|  _{p} \le \epsilon \label{con:objective}
\end{split}
\end{equation}
where $J_{MMD}(\cdot)$ represents the MMD value between the two feature distributions in a certain detector layer. We calculate the parameters of adversarial exposure, blur, and noise by optimizing Eq. \eqref{con:objective}.

\vspace{0.2em}
{\bf Algorithm.} 
As shown in Figure \ref{fig:methods}. First, we initialize the parameters $\left \{ \textbf{a},\varphi,\sigma,\mathbf{N}_{a} \right \}$ for the three perturbations and add them to the fake images. Next, we collect the feature distributions of both the natural and fake images on the global pooling layer in each DNN-based detector model and calculate their MMD values. To optimize the perturbation parameters, we use sign gradient descent to minimize Eq. \eqref{con:objective}. Finally, we apply the optimized values of $\left \{ \textbf{a},\varphi,\sigma,\mathbf{N}_{a} \right \}$ to the raw fake images, resulting in the generation of adversarial fake images.

\section{Multi-layer StatAttack (MStatAttack)}

Repeatedly adding adversarial perturbations to an image may increase the success rate of attacks. Hence, we further extend the StatAttack to a more powerful version, \ie, MStatAttack, where we extend the single-layer degradation to multi-layer degradations sequentially and use the loss to tune the combination weights jointly. 

Specifically, as depicted in Figure \ref{fig:MStatAttack}, MStatAttack employs multiple layers to process the input. At the $k$th layer, MStatAttack takes the output of the $(k-1)$th layer, which is denoted as $\mathbf{X}^\text{fake}_{p_{k-1}}$, and applies three types of adversarial degradation (\ie, exposure, blur, and noise) in parallel. 
Then, we maintain four weights (\ie, $w'_{e_{k}}$, $w'_{b_{k}}$, $w'_{n_{k}}$, $w'_{\mathbf{X}_{k}}$) to mix $\mathbf{X}^\text{fake}_{p_{k-1}}$ and its three perturbed versions, obtaining the output of the $k$th layer.
To make sure the sum of the four weights equal one, we feed them to a softmax layer and get (\ie, $w_{e_{k}}$, $w_{b_{k}}$, $w_{n_{k}}$, $w_{\mathbf{X}_{k}}$ ).
The perturbation of each layer can be written as
\begin{equation}
\begin{split}
    \mathbf{X}^\text{fake}_{p_{k}} &= w_{e_{k}}P_{e}(\mathbf{X}^\text{fake}_{p_{k-1}})+w_{b_{k}}P_{b}(\mathbf{X}^\text{fake}_{p_{k-1}})\\  &+w_{n_{k}}P_{k}(\mathbf{X}^\text{fake}_{p_{n-1}}) + w_{\mathbf{X}_{k}}\mathbf{X}^\text{fake}_{p_{k-1}} ,\ n \ge 1
\end{split}
\end{equation}
where the $\mathbf{X}^\text{fake}_{p_{k}}$ represents the adversarial examples generated at $k$th layer. 
During optimization, the raw weights and all degradation parameters are jointly optimized based on the loss function. The softmax layer ensures that the updated weights add up to one, preventing intensity overflow during the image mixing process.

In addition to automatically selecting the appropriate attack strategy ( \ie, perturbation combinations ), MStatAttack can enhances the generated adversarial examples by incorporating more components from the raw fake images, resulting in visually closer raw fake images.
\begin{table}[ht]
\centering
\caption{Statistics of the collected dataset. The first column is the fake faces, the second column is the type of fake faces, the third column is the collection method, and the fourth column is the real data source for generating the fake faces.}
\resizebox{\linewidth}{!}{%
\begin{tabular}{@{}l|l|l|l@{}}
\toprule
Fake Data & Fake Type & Collection & Real Source \\ \midrule
FF++ & face identity swap & FaceForensics++ & FaceForensics++ \\
StyleGANv2 & entire face syhthesis & self-synthesis & FFHQ \\
StarGAN & face manipulation & self-synthesis & CelebA \\
ProGAN & entire face syhthesis & self-synthesis & CelebA-HQ \\ \bottomrule
\end{tabular}%
}
\label{tab:Statistic of fake dataset}
\end{table}
\section{Experiments}
\subsection{Experimental Setups}
\textbf{Dataset.}
We conduct experiments on four datasets to evaluate our method. For entire face synthesis, we use StyleGANv2 \cite{karras2019style} and ProGAN \cite{karras2017progressive} to generate high-quality fake face images. For face attribute manipulation, we employ StarGAN \cite{choi2018stargan} to modify the attributes of natural face images, where the modified attributes are hair color, gender, and age. For face identity swap, we use the DeepFake dataset provided in the FaceForensics++ \cite{rossler2018faceforensics} benchmark, which was produced using the publicly available tool. Table~\ref{tab:Statistic of fake dataset} presents the statistics of our collected fake dataset. For real face images, we use real face images from  CelebA, FFHQ, and FaceForensics++ dataset. 

\textbf{Victim detectors.}
For the spatial-based detectors, we use four spatial-based classification models, \ie, ResNet50 \cite{he2016deep}, EfficentNet-b4 \cite{tan2019efficientnet}, DenseNet\cite{huang2017densely}, and MobileNet\cite{howard2017mobilenets}. For the frequency-based detectors, we adopt two popular detectors, DCTA\cite{frank2020leveraging} and DFTD\cite{zhang2019detecting}.  To ensure fair evaluations, we reproduce the PyTorch version of DCTA and rewrite the frequency domain transform operations involved in both methods as a network structure layer. Table~\ref{tab:acc} shows the accuracy of these detectors on each dataset.

\begin{table}[t]
\caption{The accuracy of each detector on different datasets.}
\resizebox{\linewidth}{!}{%
\begin{tabular}{@{}c|cccccc@{}}
\toprule
\begin{tabular}[c]{@{}c@{}}Model\&\\ DataSet\end{tabular}\begin{tabular}[c]{@{}c@{}}Model\&\\ DataSet\end{tabular} & ResNet & EffNet & DenNet & MobNet & DCTA & DFTD \\ \midrule
FF++   & 94.2\% & 93.8\% & 91.5\% & 94.4\% & 93.1\% & 91.6\% \\
StyleGANv2 & 98.7\% & 97.5\% & 99.2\% & 97.2\% & 98.5\% & 97.4\% \\
StarGAN    & 97.6\% & 98.9\% & 98.4\% & 96.4\% & 100\%  & 98.8\% \\
ProGAN     & 99.1\% & 99.3\% & 98.0\% & 98.9\% & 99.2\% & 99.0\% \\ \bottomrule
\end{tabular}%
}
\label{tab:acc}
\end{table}

\vspace{0.2em}
\textbf{Metrics.}
We employ the attack success rate (ASR) and image quality metric (\ie, BRISQUE~\cite{mittal2012no}) to evaluate the effectiveness of our attack methods. BRISQUE is a non-referenced image quality assessment metric, where a higher BRISQUE score indicates lower image quality.

\vspace{0.2em}
\textbf{Baseline attacks.}
We compare with two commonly used baselines attack, \ie, PGD\cite{madry2017towards} and FGSM\cite{goodfellow2014explaining}, and two SOTA transferable-based attack, \ie, MIFGSM\cite{dong2018boosting} and VMIFGSM \cite{wang2021enhancing}. For the parameters of baseline attacks, we set the max perturbation magnitude $\epsilon=8/255$ with the pixel range [0,1], and the iteration number is 40.

\begin{table*}[ht]
\centering
\caption{The attack success rate and image quality assessment on spatial-based Deepfake detectors. In each group, the first column denotes the attack success rates in the white-box setting, while the second, third and fourth columns are the 
transfer attack success rates on other Deepfake detectors. We mark the first, second, and third highest attack success rates in red, yellow, and blue. The last column shows the BRISQUE score. }
\resizebox{\textwidth}{!}{%
\begin{tabular}{@{}lc|ccccc|ccccc|ccccc|ccccc@{}}
\toprule
 & \cellcolor[HTML]{EFEFEF}{\color[HTML]{333333} Crafted from} & \multicolumn{1}{l}{\cellcolor[HTML]{EFEFEF}} & \multicolumn{3}{c}{\cellcolor[HTML]{EFEFEF}{\color[HTML]{333333} ResNet}} & \multicolumn{1}{l|}{\cellcolor[HTML]{EFEFEF}} & \multicolumn{1}{l}{\cellcolor[HTML]{EFEFEF}} & \multicolumn{3}{c}{\cellcolor[HTML]{EFEFEF}{\color[HTML]{333333} EfficinetNet}} & \multicolumn{1}{l|}{\cellcolor[HTML]{EFEFEF}} & \multicolumn{1}{l}{\cellcolor[HTML]{EFEFEF}} & \multicolumn{3}{c}{\cellcolor[HTML]{EFEFEF}{\color[HTML]{333333} DenseNet}} & \multicolumn{1}{l|}{\cellcolor[HTML]{EFEFEF}} & \multicolumn{1}{l}{\cellcolor[HTML]{EFEFEF}} & \multicolumn{3}{c}{\cellcolor[HTML]{EFEFEF}{\color[HTML]{333333} MobileNet}} & \multicolumn{1}{l}{\cellcolor[HTML]{EFEFEF}} \\ \midrule
\multicolumn{1}{c|}{} & \cellcolor[HTML]{EFEFEF}{\color[HTML]{333333} \begin{tabular}[c]{@{}c@{}}Attack model \\ \& Metrics\end{tabular}} & \multicolumn{1}{l|}{\cellcolor[HTML]{EFEFEF}ResNet} & \cellcolor[HTML]{EFEFEF}{\color[HTML]{333333} EffNet} & \cellcolor[HTML]{EFEFEF}{\color[HTML]{333333} DenNet} & \cellcolor[HTML]{EFEFEF}{\color[HTML]{333333} MobNet} & \multicolumn{1}{l|}{\cellcolor[HTML]{EFEFEF}BRISQUE} & \multicolumn{1}{l|}{\cellcolor[HTML]{EFEFEF}EffNet} & \cellcolor[HTML]{EFEFEF}{\color[HTML]{333333} ResNet} & \cellcolor[HTML]{EFEFEF}{\color[HTML]{333333} DenNet} & \cellcolor[HTML]{EFEFEF}{\color[HTML]{333333} MobNet} & \multicolumn{1}{l|}{\cellcolor[HTML]{EFEFEF}BRISQUE} & \multicolumn{1}{l|}{\cellcolor[HTML]{EFEFEF}DenseNet} & \cellcolor[HTML]{EFEFEF}{\color[HTML]{333333} ResNet} & \cellcolor[HTML]{EFEFEF}{\color[HTML]{333333} EffNet} & \cellcolor[HTML]{EFEFEF}{\color[HTML]{333333} MobNet} & \multicolumn{1}{l|}{\cellcolor[HTML]{EFEFEF}BRISQUE} & \multicolumn{1}{l|}{\cellcolor[HTML]{EFEFEF}MobNet} & \cellcolor[HTML]{EFEFEF}{\color[HTML]{333333} ResNet} & \cellcolor[HTML]{EFEFEF}{\color[HTML]{333333} EffNet} & \cellcolor[HTML]{EFEFEF}{\color[HTML]{333333} DenNet} & \multicolumn{1}{l}{\cellcolor[HTML]{EFEFEF}BRISQUE} \\ \midrule
\multicolumn{1}{c|}{{\color[HTML]{333333} }} & {\color[HTML]{333333} PGD} & \multicolumn{1}{c|}{{\color[HTML]{333333} 78.3\%}} & {\color[HTML]{333333} 54.7\%} & {\color[HTML]{333333} 53.1\%} & {\color[HTML]{333333} 60.0\%} & {\color[HTML]{333333} 49.4} & \multicolumn{1}{c|}{{\color[HTML]{333333} 96.1\%}} & {\color[HTML]{333333} 16.5\%} & {\color[HTML]{333333} 13.3\%} & {\color[HTML]{333333} 43.3\%} & {\color[HTML]{333333} 48.2} & \multicolumn{1}{c|}{{\color[HTML]{333333} 95.7\%}} & {\color[HTML]{333333} 43.3\%} & {\color[HTML]{333333} 38.1\%} & {\color[HTML]{333333} 59.4\%} & {\color[HTML]{333333} 46.5} & \multicolumn{1}{c|}{{\color[HTML]{333333} 94\%}} & {\color[HTML]{333333} 10.0\%} & {\color[HTML]{333333} 25.9\%} & {\color[HTML]{333333} 6.0\%} & {\color[HTML]{333333} 50.1} \\
\multicolumn{1}{c|}{{\color[HTML]{333333} }} & {\color[HTML]{333333} FGSM} & \multicolumn{1}{c|}{{\color[HTML]{333333} 88.8\%}} & {\color[HTML]{333333} 31.4\%} & {\color[HTML]{333333} 30.5\%} & {\color[HTML]{333333} 44.3\%} & {\color[HTML]{333333} 52.3} & \multicolumn{1}{c|}{{\color[HTML]{333333} 84.6\%}} & \cellcolor[HTML]{96FFFB}51.5\% & {\color[HTML]{333333} 44.7\%} & {\color[HTML]{333333} 41.5\%} & {\color[HTML]{333333} 51.3} & \multicolumn{1}{c|}{{\color[HTML]{333333} 91.6\%}} & {\color[HTML]{333333} 59.8\%} & {\color[HTML]{333333} 44.8\%} & {\color[HTML]{333333} 48.5\%} & {\color[HTML]{333333} 49.6} & \multicolumn{1}{c|}{{\color[HTML]{333333} 93.5\%}} & {\color[HTML]{333333} 36.8\%} & {\color[HTML]{333333} 42.5\%} & {\color[HTML]{333333} 33.6\%} & {\color[HTML]{333333} 52.7} \\
\multicolumn{1}{c|}{{\color[HTML]{333333} }} & {\color[HTML]{333333} MIFGSM} & \multicolumn{1}{c|}{{\color[HTML]{333333} 98.2\%}} & 47.2\% & 45.5\% & 63.2\% & 46.3 & \multicolumn{1}{c|}{{\color[HTML]{333333} 97.2\%}} & 45.2\% & 47.6\% & 51.3\% & {\color[HTML]{333333} 44.2} & \multicolumn{1}{c|}{{\color[HTML]{333333} 100\%}} & 68.5\% & 57.4\% & 52.3\% & {\color[HTML]{333333} 43.8} & \multicolumn{1}{c|}{{\color[HTML]{333333} 100\%}} & 39.4\% & 40.2\% & 25.3\% & {\color[HTML]{333333} 48.7} \\
\multicolumn{1}{c|}{{\color[HTML]{333333} }} & {\color[HTML]{333333} VMIFGSM} & \multicolumn{1}{c|}{{\color[HTML]{333333} 98.0\%}} & \cellcolor[HTML]{96FFFB}{\color[HTML]{333333} 53.0\%} & \cellcolor[HTML]{96FFFB}{\color[HTML]{333333} 55.0\%} & \cellcolor[HTML]{96FFFB}65.3\% & 46.5 & \multicolumn{1}{c|}{{\color[HTML]{333333} 97.3\%}} & {\color[HTML]{333333} 43.9\%} & \cellcolor[HTML]{96FFFB}{\color[HTML]{333333} 49.5\%} & \cellcolor[HTML]{96FFFB}{\color[HTML]{333333} 56.3\%} & {\color[HTML]{333333} 45.4} & \multicolumn{1}{c|}{{\color[HTML]{333333} 99.7\%}} & \cellcolor[HTML]{96FFFB}72.0\% & \cellcolor[HTML]{96FFFB}58.0\% & \cellcolor[HTML]{FFCCC9}73.4\% & {\color[HTML]{333333} 44.2} & \multicolumn{1}{c|}{{\color[HTML]{333333} 97.5\%}} & \cellcolor[HTML]{96FFFB}40.9\% & \cellcolor[HTML]{96FFFB}44.3\% & \cellcolor[HTML]{96FFFB}33.9\% & {\color[HTML]{333333} 46.4} \\ \cmidrule(l){2-22} 
\multicolumn{1}{c|}{{\color[HTML]{333333} }} & {\color[HTML]{333333} \textbf{StatAttack}} & \multicolumn{1}{c|}{{\color[HTML]{333333} 96.5\%}} & \cellcolor[HTML]{FFCCC9}{\color[HTML]{333333} 65.9\%} & \cellcolor[HTML]{FFFFC7}{\color[HTML]{333333} 81.3\%} & \cellcolor[HTML]{FFFFC7}{\color[HTML]{333333} 71.2\%} & 49.5 & \multicolumn{1}{c|}{{\color[HTML]{333333} 97.6\%}} & \cellcolor[HTML]{FFFFC7}{\color[HTML]{333333} 53.2\%} & \cellcolor[HTML]{FFFFC7}{\color[HTML]{333333} 51.4\%} & \cellcolor[HTML]{FFFFC7}{\color[HTML]{333333} 66\%} & {\color[HTML]{333333} 47.2} & \multicolumn{1}{c|}{{\color[HTML]{333333} 97.3\%}} & \cellcolor[HTML]{FFFFC7}{\color[HTML]{333333} 75.1\%} & \cellcolor[HTML]{FFFFC7}{\color[HTML]{333333} 60.1\%} & \cellcolor[HTML]{96FFFB}{\color[HTML]{333333} 60.6\%} & {\color[HTML]{333333} 47.5} & \multicolumn{1}{c|}{{\color[HTML]{333333} 98.1\%}} & \cellcolor[HTML]{FFFFC7}{\color[HTML]{333333} 51.9\%} & \cellcolor[HTML]{FFFFC7}{\color[HTML]{333333} 58.1\%} & \cellcolor[HTML]{FFCCC9}{\color[HTML]{333333} 40.7\%} & {\color[HTML]{333333} 49.6} \\
\multicolumn{1}{c|}{\multirow{-6}{*}{{\color[HTML]{333333} \rotatebox{90}{FF++}}}} & {\color[HTML]{333333} \textbf{MStatAttack}} & \multicolumn{1}{c|}{{\color[HTML]{333333} 97.3\%}} & \cellcolor[HTML]{FFFFC7}{\color[HTML]{333333} 62.1\%} & \cellcolor[HTML]{FFCCC9}{\color[HTML]{333333} 83.1\%} & \cellcolor[HTML]{FFCCC9}{\color[HTML]{333333} 73.2\%} & 46.7 & \multicolumn{1}{c|}{{\color[HTML]{333333} 98.8\%}} & \cellcolor[HTML]{FFCCC9}{\color[HTML]{333333} 57.0\%} & \cellcolor[HTML]{FFCCC9}{\color[HTML]{333333} 58.8\%} & \cellcolor[HTML]{FFCCC9}{\color[HTML]{333333} 68.3\%} & {\color[HTML]{333333} 46.3} & \multicolumn{1}{c|}{{\color[HTML]{333333} 98.8\%}} & \cellcolor[HTML]{FFCCC9}{\color[HTML]{333333} 80.2\%} & \cellcolor[HTML]{FFCCC9}{\color[HTML]{333333} 70.2\%} & \cellcolor[HTML]{FFFFC7}{\color[HTML]{333333} 65.2\%} & {\color[HTML]{333333} 44.6} & \multicolumn{1}{c|}{{\color[HTML]{333333} 99.3\%}} & \cellcolor[HTML]{FFCCC9}{\color[HTML]{333333} 60.8\%} & \cellcolor[HTML]{FFCCC9}{\color[HTML]{333333} 60.1\%} & \cellcolor[HTML]{FFFFC7}{\color[HTML]{333333} 40.3\%} & {\color[HTML]{333333} 45.5} \\ \midrule
\multicolumn{1}{l|}{{\color[HTML]{333333} }} & {\color[HTML]{333333} PGD} & \multicolumn{1}{c|}{{\color[HTML]{333333} 98.2\%}} & {\color[HTML]{333333} 23.8\%} & {\color[HTML]{333333} 27.7\%} & {\color[HTML]{333333} 36.6\%} & {\color[HTML]{333333} 17.3} & \multicolumn{1}{c|}{{\color[HTML]{333333} 96.2\%}} & {\color[HTML]{333333} 10.1\%} & {\color[HTML]{333333} 12.9\%} & {\color[HTML]{333333} 17.2\%} & {\color[HTML]{333333} 18.4} & \multicolumn{1}{c|}{{\color[HTML]{333333} 96.5\%}} & {\color[HTML]{333333} 17.0\%} & {\color[HTML]{333333} 13.1\%} & {\color[HTML]{333333} 27.6\%} & {\color[HTML]{333333} 19.7} & \multicolumn{1}{c|}{{\color[HTML]{333333} 100\%}} & {\color[HTML]{333333} 8.8\%} & {\color[HTML]{333333} 12.2\%} & {\color[HTML]{333333} 10.5\%} & {\color[HTML]{333333} 18.6} \\
\multicolumn{1}{l|}{{\color[HTML]{333333} }} & {\color[HTML]{333333} FGSM} & \multicolumn{1}{c|}{{\color[HTML]{333333} 78.3\%}} & {\color[HTML]{333333} 11.4\%} & {\color[HTML]{333333} 18.7\%} & {\color[HTML]{333333} 34.4\%} & {\color[HTML]{333333} 21.3} & \multicolumn{1}{c|}{{\color[HTML]{333333} 89.0\%}} & 8.0\% & 5.9\% & 9.5\% & {\color[HTML]{333333} 26.5} & \multicolumn{1}{c|}{{\color[HTML]{333333} 75.4\%}} & 15.6\% & 8.1\% & 36.1\% & {\color[HTML]{333333} 28.8} & \multicolumn{1}{c|}{{\color[HTML]{333333} 85.6\%}} & {\color[HTML]{333333} 10.4\%} & {\color[HTML]{333333} 6.5\%} & {\color[HTML]{333333} 8.2\%} & {\color[HTML]{333333} 26.4} \\
\multicolumn{1}{l|}{{\color[HTML]{333333} }} & {\color[HTML]{333333} MIFGSM} & \multicolumn{1}{c|}{{\color[HTML]{333333} 99.5\%}} & \cellcolor[HTML]{96FFFB}{\color[HTML]{333333} 24.3\%} & \cellcolor[HTML]{96FFFB}{\color[HTML]{333333} 29.1\%} & \cellcolor[HTML]{96FFFB}{\color[HTML]{333333} 38.9\%} & {\color[HTML]{333333} 15.3} & \multicolumn{1}{c|}{{\color[HTML]{333333} 98.5\%}} & \cellcolor[HTML]{96FFFB}10.3\% & 18.2\% & 17.5\% & {\color[HTML]{333333} 16.3} & \multicolumn{1}{c|}{{\color[HTML]{333333} 98.2\%}} & 18.5\% & 10.2\% & 35.2\% & {\color[HTML]{333333} 17.5} & \multicolumn{1}{c|}{{\color[HTML]{333333} 100\%}} & \cellcolor[HTML]{96FFFB}{\color[HTML]{333333} 18.2\%} & \cellcolor[HTML]{96FFFB}{\color[HTML]{333333} 13.2\%} & \cellcolor[HTML]{96FFFB}{\color[HTML]{333333} 24.4\%} & {\color[HTML]{333333} 17.1} \\
\multicolumn{1}{l|}{{\color[HTML]{333333} }} & {\color[HTML]{333333} VMIFGSM} & \multicolumn{1}{c|}{{\color[HTML]{333333} 100\%}} & {\color[HTML]{333333} 21.4\%} & {\color[HTML]{333333} 28.7\%} & {\color[HTML]{333333} 36.1\%} & {\color[HTML]{333333} 16.4} & \multicolumn{1}{c|}{{\color[HTML]{333333} 96.8\%}} & 9.7\% & \cellcolor[HTML]{96FFFB}{\color[HTML]{333333} 19.8\%} & \cellcolor[HTML]{96FFFB}{\color[HTML]{333333} 18.3\%} & {\color[HTML]{333333} 17.2} & \multicolumn{1}{c|}{{\color[HTML]{333333} 98.4\%}} & \cellcolor[HTML]{96FFFB}{\color[HTML]{333333} 20.3\%} & \cellcolor[HTML]{96FFFB}{\color[HTML]{333333} 15.7\%} & \cellcolor[HTML]{96FFFB}{\color[HTML]{333333} 41.5\%} & {\color[HTML]{333333} 19.3} & \multicolumn{1}{c|}{{\color[HTML]{333333} 100\%}} & {\color[HTML]{333333} 17.3\%} & {\color[HTML]{333333} 11.1\%} & {\color[HTML]{333333} 13.1\%} & {\color[HTML]{333333} 18.8} \\ \cmidrule(l){2-22} 
\multicolumn{1}{l|}{{\color[HTML]{333333} }} & {\color[HTML]{333333} \textbf{StatAttack}} & \multicolumn{1}{c|}{{\color[HTML]{333333} 98.3\%}} & \cellcolor[HTML]{FFFFC7}{\color[HTML]{333333} 43.2\%} & \cellcolor[HTML]{FFFFC7}{\color[HTML]{333333} 57.1\%} & \cellcolor[HTML]{FFFFC7}{\color[HTML]{333333} 64.5\%} & {\color[HTML]{333333} 17.5} & \multicolumn{1}{c|}{{\color[HTML]{333333} 96.6\%}} & \cellcolor[HTML]{FFFFC7}{\color[HTML]{333333} 33.6\%} & \cellcolor[HTML]{FFCCC9}{\color[HTML]{333333} 73.6\%} & \cellcolor[HTML]{FFCCC9}{\color[HTML]{333333} 78.2\%} & {\color[HTML]{333333} 18.3} & \multicolumn{1}{c|}{{\color[HTML]{333333} 97.1\%}} & \cellcolor[HTML]{FFFFC7}{\color[HTML]{333333} 46.6\%} & \cellcolor[HTML]{FFFFC7}{\color[HTML]{333333} 31.7\%} & \cellcolor[HTML]{FFFFC7}{\color[HTML]{333333} 77.4\%} & {\color[HTML]{333333} 18.5} & \multicolumn{1}{c|}{{\color[HTML]{333333} 98.0\%}} & \cellcolor[HTML]{FFFFC7}{\color[HTML]{333333} 26.5\%} & \cellcolor[HTML]{FFFFC7}{\color[HTML]{333333} 25.5\%} & \cellcolor[HTML]{FFCCC9}{\color[HTML]{333333} 60\%} & {\color[HTML]{333333} 17.5} \\
\multicolumn{1}{l|}{\multirow{-6}{*}{{\color[HTML]{333333} \rotatebox{90}{StyleGANv2}}}} & {\color[HTML]{333333} \textbf{MStatAttack}} & \multicolumn{1}{c|}{{\color[HTML]{333333} 100\%}} & \cellcolor[HTML]{FFCCC9}{\color[HTML]{333333} 46.2\%} & \cellcolor[HTML]{FFCCC9}{\color[HTML]{333333} 62.1\%} & \cellcolor[HTML]{FFCCC9}{\color[HTML]{333333} 68.2\%} & {\color[HTML]{333333} 16.8} & \multicolumn{1}{c|}{{\color[HTML]{333333} 97.9\%}} & \cellcolor[HTML]{FFCCC9}{\color[HTML]{333333} 41.2\%} & \cellcolor[HTML]{FFFFC7}{\color[HTML]{333333} 75.8\%} & \cellcolor[HTML]{FFFFC7}{\color[HTML]{333333} 76.6\%} & {\color[HTML]{333333} 17.5} & \multicolumn{1}{c|}{{\color[HTML]{333333} 98.6\%}} & \cellcolor[HTML]{FFCCC9}{\color[HTML]{333333} 47.2\%} & \cellcolor[HTML]{FFCCC9}{\color[HTML]{333333} 34.5\%} & \cellcolor[HTML]{FFCCC9}{\color[HTML]{333333} 78.0\%} & {\color[HTML]{333333} 16.4} & \multicolumn{1}{c|}{{\color[HTML]{333333} 98.9\%}} & \cellcolor[HTML]{FFCCC9}{\color[HTML]{333333} 37.6\%} & \cellcolor[HTML]{FFCCC9}{\color[HTML]{333333} 30.5\%} & \cellcolor[HTML]{FFFFC7}{\color[HTML]{333333} 58\%} & {\color[HTML]{333333} 17.1} \\ \midrule
\multicolumn{1}{l|}{{\color[HTML]{333333} }} & {\color[HTML]{333333} PGD} & \multicolumn{1}{c|}{{\color[HTML]{333333} 96\%}} & {\color[HTML]{333333} 13.9\%} & {\color[HTML]{333333} 20.5\%} & {\color[HTML]{333333} 20.5\%} & {\color[HTML]{333333} 32.9} & \multicolumn{1}{c|}{{\color[HTML]{333333} 88.0\%}} & {\color[HTML]{333333} 30.1\%} & {\color[HTML]{333333} 22.2\%} & {\color[HTML]{333333} 18.4\%} & {\color[HTML]{333333} 30.2} & \multicolumn{1}{c|}{{\color[HTML]{333333} 92.8\%}} & {\color[HTML]{333333} 68.4\%} & {\color[HTML]{333333} 32.9\%} & {\color[HTML]{333333} 24.8\%} & {\color[HTML]{333333} 33.1} & \multicolumn{1}{c|}{{\color[HTML]{333333} 100\%}} & {\color[HTML]{333333} 8.8\%} & {\color[HTML]{333333} 10.2\%} & {\color[HTML]{333333} 6.5\%} & {\color[HTML]{333333} 30.9} \\
\multicolumn{1}{l|}{{\color[HTML]{333333} }} & {\color[HTML]{333333} FGSM} & \multicolumn{1}{c|}{{\color[HTML]{333333} 89.8\%}} & {\color[HTML]{333333} 32.7\%} & {\color[HTML]{333333} 56.4\%} & {\color[HTML]{333333} 36.0\%} & {\color[HTML]{333333} 40.6} & \multicolumn{1}{c|}{{\color[HTML]{333333} 75.5\%}} & {\color[HTML]{333333} 33.8\%} & {\color[HTML]{333333} 44.5\%} & {\color[HTML]{333333} 44.9\%} & {\color[HTML]{333333} 38.8} & \multicolumn{1}{c|}{{\color[HTML]{333333} 82.1\%}} & {\color[HTML]{333333} 65.8\%} & {\color[HTML]{333333} 26.2\%} & {\color[HTML]{333333} 32.2\%} & {\color[HTML]{333333} 36.9} & \multicolumn{1}{c|}{{\color[HTML]{333333} 86.3\%}} & {\color[HTML]{333333} 52.1\%} & {\color[HTML]{333333} 18.3\%} & {\color[HTML]{333333} 16.5\%} & {\color[HTML]{333333} 37.1} \\
\multicolumn{1}{l|}{{\color[HTML]{333333} }} & {\color[HTML]{333333} MIFGSM} & \multicolumn{1}{c|}{{\color[HTML]{333333} 100\%}} & \multicolumn{1}{l}{{\color[HTML]{333333} 35.4\%}} & \multicolumn{1}{l}{{\color[HTML]{333333} 55.2\%}} & \multicolumn{1}{l}{{\color[HTML]{333333} 25.1\%}} & {\color[HTML]{333333} 30.2} & \multicolumn{1}{c|}{{\color[HTML]{333333} 99.3\%}} & \multicolumn{1}{l}{{\color[HTML]{333333} 40.0\%}} & \multicolumn{1}{l}{{\color[HTML]{333333} 57.1\%}} & \multicolumn{1}{l}{{\color[HTML]{333333} 50.2\%}} & {\color[HTML]{333333} 31.5} & \multicolumn{1}{c|}{{\color[HTML]{333333} 97.5\%}} & \multicolumn{1}{l}{{\color[HTML]{333333} 70.0\%}} & \multicolumn{1}{l}{\cellcolor[HTML]{FFCCC9}{\color[HTML]{333333} 65.5\%}} & \multicolumn{1}{l}{{\color[HTML]{333333} 33.5\%}} & {\color[HTML]{333333} 30.4} & \multicolumn{1}{c|}{{\color[HTML]{333333} 100\%}} & \multicolumn{1}{l}{{\color[HTML]{333333} 52.1\%}} & \cellcolor[HTML]{96FFFB}{\color[HTML]{333333} 39.5\%} & \cellcolor[HTML]{96FFFB}{\color[HTML]{333333} 55.8\%} & {\color[HTML]{333333} 30.3} \\
\multicolumn{1}{l|}{{\color[HTML]{333333} }} & {\color[HTML]{333333} VMIFGSM} & \multicolumn{1}{c|}{{\color[HTML]{333333} 100\%}} & \cellcolor[HTML]{96FFFB}{\color[HTML]{333333} 39.8\%} & \cellcolor[HTML]{96FFFB}{\color[HTML]{333333} 75.4\%} & \cellcolor[HTML]{96FFFB}{\color[HTML]{333333} 36.8\%} & {\color[HTML]{333333} 31.5} & \multicolumn{1}{c|}{{\color[HTML]{333333} 99.7\%}} & \cellcolor[HTML]{96FFFB}{\color[HTML]{333333} 59.9\%} & \cellcolor[HTML]{96FFFB}{\color[HTML]{333333} 58.3\%} & \cellcolor[HTML]{96FFFB}{\color[HTML]{333333} 55.2\%} & {\color[HTML]{333333} 32.8} & \multicolumn{1}{c|}{{\color[HTML]{333333} 97.8\%}} & \cellcolor[HTML]{96FFFB}{\color[HTML]{333333} 72.2\%} & 57.7\% & \cellcolor[HTML]{96FFFB}{\color[HTML]{333333} 56.7\%} & {\color[HTML]{333333} 33.7} & \multicolumn{1}{c|}{{\color[HTML]{333333} 100\%}} & \cellcolor[HTML]{96FFFB}{\color[HTML]{333333} 53.2\%} & 24.1\% & 36.5\% & {\color[HTML]{333333} 30.5} \\ \cmidrule(l){2-22} 
\multicolumn{1}{l|}{{\color[HTML]{333333} }} & {\color[HTML]{333333} \textbf{StatAttack}} & \multicolumn{1}{c|}{{\color[HTML]{333333} 99.3\%}} & \cellcolor[HTML]{FFFFC7}{\color[HTML]{333333} 41.0\%} & \cellcolor[HTML]{FFFFC7}{\color[HTML]{333333} 88.4\%} & \cellcolor[HTML]{FFFFC7}{\color[HTML]{333333} 39.1\%} & {\color[HTML]{333333} 32.5} & \multicolumn{1}{c|}{{\color[HTML]{333333} 95.7\%}} & \cellcolor[HTML]{FFFFC7}{\color[HTML]{333333} 68.8\%} & \cellcolor[HTML]{FFFFC7}{\color[HTML]{333333} 61.6\%} & \cellcolor[HTML]{FFFFC7}{\color[HTML]{333333} 62.0\%} & {\color[HTML]{333333} 30.4} & \multicolumn{1}{c|}{{\color[HTML]{333333} 97.9\%}} & \cellcolor[HTML]{FFFFC7}{\color[HTML]{333333} 78.1\%} & \cellcolor[HTML]{FFFFC7}{\color[HTML]{333333} 64.3\%} & \cellcolor[HTML]{FFFFC7}{\color[HTML]{333333} 63.3\%} & {\color[HTML]{333333} 31.2} & \multicolumn{1}{c|}{{\color[HTML]{333333} 100\%}} & \cellcolor[HTML]{FFFFC7}{\color[HTML]{333333} 82.5\%} & \cellcolor[HTML]{FFFFC7}{\color[HTML]{333333} 42.1\%} & \cellcolor[HTML]{FFFFC7}{\color[HTML]{333333} 75.1\%} & {\color[HTML]{333333} 30.4} \\
\multicolumn{1}{l|}{\multirow{-6}{*}{{\color[HTML]{333333} \rotatebox{90}{StarGAN}}}} & {\color[HTML]{333333} \textbf{MStatAttack}} & \multicolumn{1}{c|}{{\color[HTML]{333333} 100\%}} & \cellcolor[HTML]{FFCCC9}{\color[HTML]{333333} 48.0\%} & \cellcolor[HTML]{FFCCC9}{\color[HTML]{333333} 89.6\%} & \cellcolor[HTML]{FFCCC9}{\color[HTML]{333333} 42.1\%} & {\color[HTML]{333333} 30.6} & \multicolumn{1}{c|}{{\color[HTML]{333333} 97.6\%}} & \cellcolor[HTML]{FFCCC9}{\color[HTML]{333333} 75.6\%} & \cellcolor[HTML]{FFCCC9}{\color[HTML]{333333} 62.3\%} & \cellcolor[HTML]{FFCCC9}{\color[HTML]{333333} 68.3\%} & {\color[HTML]{333333} 31.6} & \multicolumn{1}{c|}{{\color[HTML]{333333} 98.6\%}} & \cellcolor[HTML]{FFCCC9}{\color[HTML]{333333} 88.3\%} & \cellcolor[HTML]{96FFFB}63.1\% & \cellcolor[HTML]{FFCCC9}{\color[HTML]{333333} 71.3\%} & {\color[HTML]{333333} 29.9} & \multicolumn{1}{c|}{{\color[HTML]{333333} 99.3\%}} & \cellcolor[HTML]{FFCCC9}{\color[HTML]{333333} 87.2\%} & \cellcolor[HTML]{FFCCC9}{\color[HTML]{333333} 43.5\%} & \cellcolor[HTML]{FFCCC9}{\color[HTML]{333333} 77.9\%} & {\color[HTML]{333333} 28.8} \\ \midrule
\multicolumn{1}{l|}{{\color[HTML]{333333} }} & {\color[HTML]{333333} PGD} & \multicolumn{1}{c|}{{\color[HTML]{333333} 93.8\%}} & {\color[HTML]{333333} 13.2\%} & {\color[HTML]{333333} 20.1\%} & {\color[HTML]{333333} 10.1\%} & {\color[HTML]{333333} 18.8} & \multicolumn{1}{c|}{{\color[HTML]{333333} 92\%}} & {\color[HTML]{333333} 23.5\%} & {\color[HTML]{333333} 21.2\%} & 34.5\% & {\color[HTML]{333333} 15.1} & \multicolumn{1}{c|}{{\color[HTML]{333333} 96.5\%}} & {\color[HTML]{333333} 7.2\%} & {\color[HTML]{333333} 16.5\%} & {\color[HTML]{333333} 25.8\%} & {\color[HTML]{333333} 15.6} & \multicolumn{1}{c|}{{\color[HTML]{333333} 97.3\%}} & {\color[HTML]{333333} 32.8\%} & {\color[HTML]{333333} 21.4\%} & {\color[HTML]{333333} 33.1\%} & {\color[HTML]{333333} 16.1} \\
\multicolumn{1}{l|}{{\color[HTML]{333333} }} & {\color[HTML]{333333} FGSM} & \multicolumn{1}{c|}{{\color[HTML]{333333} 76.1\%}} & {\color[HTML]{333333} 20.5\%} & {\color[HTML]{333333} 20.1\%} & {\color[HTML]{333333} 30.4\%} & {\color[HTML]{333333} 21.2} & \multicolumn{1}{c|}{{\color[HTML]{333333} 78.6\%}} & \multicolumn{1}{l}{{\color[HTML]{333333} 30.2\%}} & \multicolumn{1}{l}{{\color[HTML]{333333} 35.7\%}} & \multicolumn{1}{l}{{\color[HTML]{333333} 35.4\%}} & {\color[HTML]{333333} 20.4} & \multicolumn{1}{c|}{{\color[HTML]{333333} 81.3\%}} & \multicolumn{1}{l}{{\color[HTML]{333333} 18.7\%}} & \multicolumn{1}{l}{{\color[HTML]{333333} 30.1\%}} & {\color[HTML]{333333} 34.2\%} & {\color[HTML]{333333} 19.6} & \multicolumn{1}{c|}{{\color[HTML]{333333} 92.8\%}} & \multicolumn{1}{l}{34.5\%} & \multicolumn{1}{l}{{\color[HTML]{333333} 30.2\%}} & \multicolumn{1}{l}{{\color[HTML]{333333} 38.8\%}} & {\color[HTML]{333333} 21.4} \\
\multicolumn{1}{l|}{{\color[HTML]{333333} }} & {\color[HTML]{333333} MIFGSM} & \multicolumn{1}{c|}{{\color[HTML]{333333} 96.5\%}} & {\color[HTML]{333333} 20.2\%} & {\color[HTML]{333333} 13.8\%} & 31.6\% & {\color[HTML]{333333} 18.5} & \multicolumn{1}{c|}{{\color[HTML]{333333} 99.5\%}} & {\color[HTML]{333333} 22.5\%} & {\color[HTML]{333333} 28.6\%} & {\color[HTML]{333333} 30.1\%} & {\color[HTML]{333333} 18.9} & \multicolumn{1}{c|}{{\color[HTML]{333333} 97.8\%}} & {\color[HTML]{333333} 11.1\%} & {\color[HTML]{333333} 23.2\%} & \cellcolor[HTML]{96FFFB}{\color[HTML]{333333} 41.5\%} & {\color[HTML]{333333} 20.1} & \multicolumn{1}{c|}{{\color[HTML]{333333} 97.5\%}} & {\color[HTML]{333333} 37.9\%} & {\color[HTML]{333333} 22.9\%} & {\color[HTML]{333333} 23.4\%} & {\color[HTML]{333333} 19.5} \\
\multicolumn{1}{l|}{{\color[HTML]{333333} }} & {\color[HTML]{333333} VMIFGSM} & \multicolumn{1}{c|}{{\color[HTML]{333333} 97.2\%}} & \cellcolor[HTML]{96FFFB}{\color[HTML]{333333} 27.5\%} & \cellcolor[HTML]{96FFFB}{\color[HTML]{333333} 23.3\%} & \cellcolor[HTML]{96FFFB}32.4\% & {\color[HTML]{333333} 17.4} & \multicolumn{1}{c|}{{\color[HTML]{333333} 98.4\%}} & \cellcolor[HTML]{96FFFB}{\color[HTML]{333333} 37.2\%} & \cellcolor[HTML]{96FFFB}{\color[HTML]{333333} 37.7\%} & \cellcolor[HTML]{96FFFB}{\color[HTML]{333333} 37.9\%} & {\color[HTML]{333333} 16.5} & \multicolumn{1}{c|}{{\color[HTML]{333333} 96.2\%}} & \cellcolor[HTML]{96FFFB}{\color[HTML]{333333} 20.1\%} & \cellcolor[HTML]{96FFFB}{\color[HTML]{333333} 32.2\%} & {\color[HTML]{333333} 36.6\%} & {\color[HTML]{333333} 18.4} & \multicolumn{1}{c|}{{\color[HTML]{333333} 95.1\%}} & \cellcolor[HTML]{96FFFB}47.5\% & \cellcolor[HTML]{96FFFB}{\color[HTML]{333333} 32.1\%} & \cellcolor[HTML]{96FFFB}{\color[HTML]{333333} 43.5\%} & {\color[HTML]{333333} 17.1} \\ \cmidrule(l){2-22} 
\multicolumn{1}{l|}{{\color[HTML]{333333} }} & {\color[HTML]{333333} \textbf{StatAttack}} & \multicolumn{1}{c|}{{\color[HTML]{333333} 96.8\%}} & \cellcolor[HTML]{FFCCC9}{\color[HTML]{333333} 72.8\%} & \cellcolor[HTML]{FFCCC9}{\color[HTML]{333333} 78.4\%} & \cellcolor[HTML]{FFFFC7}{\color[HTML]{333333} 50.6\%} & {\color[HTML]{333333} 18.8} & \multicolumn{1}{c|}{{\color[HTML]{333333} 97.3\%}} & \cellcolor[HTML]{FFFFC7}{\color[HTML]{333333} 71.0\%} & \cellcolor[HTML]{FFFFC7}{\color[HTML]{333333} 68.5\%} & \cellcolor[HTML]{FFFFC7}{\color[HTML]{333333} 53.7\%} & {\color[HTML]{333333} 19.1} & \multicolumn{1}{c|}{{\color[HTML]{333333} 98.8\%}} & \cellcolor[HTML]{FFCCC9}{\color[HTML]{333333} 85\%} & \cellcolor[HTML]{FFCCC9}{\color[HTML]{333333} 67.2\%} & \cellcolor[HTML]{FFCCC9}{\color[HTML]{333333} 78.2\%} & {\color[HTML]{333333} 17.2} & \multicolumn{1}{c|}{{\color[HTML]{333333} 95.7\%}} & \cellcolor[HTML]{FFFFC7}{\color[HTML]{333333} 69.1\%} & \cellcolor[HTML]{FFFFC7}{\color[HTML]{333333} 62.6\%} & \cellcolor[HTML]{FFFFC7}{\color[HTML]{333333} 61.4\%} & {\color[HTML]{333333} 19.6} \\
\multicolumn{1}{l|}{\multirow{-6}{*}{{\color[HTML]{333333} \rotatebox{90}{ProGAN}}}} & {\color[HTML]{333333} \textbf{MStatAttack}} & \multicolumn{1}{c|}{{\color[HTML]{333333} 98.8\%}} & \cellcolor[HTML]{FFFFC7}{\color[HTML]{333333} 68.8\%} & \cellcolor[HTML]{FFFFC7}{\color[HTML]{333333} 75.6\%} & \cellcolor[HTML]{FFCCC9}{\color[HTML]{333333} 51.2\%} & {\color[HTML]{333333} 17.1} & \multicolumn{1}{c|}{{\color[HTML]{333333} 98.6\%}} & \cellcolor[HTML]{FFCCC9}{\color[HTML]{333333} 86.3\%} & \cellcolor[HTML]{FFCCC9}{\color[HTML]{333333} 76.0\%} & \cellcolor[HTML]{FFCCC9}{\color[HTML]{333333} 60.2\%} & {\color[HTML]{333333} 17.0} & \multicolumn{1}{c|}{{\color[HTML]{333333} 99.1\%}} & \cellcolor[HTML]{FFFFC7}{\color[HTML]{333333} 79.6\%} & \cellcolor[HTML]{FFFFC7}{\color[HTML]{333333} 65.5\%} & \cellcolor[HTML]{FFFFC7}{\color[HTML]{333333} 76.5\%} & 17.3 & \multicolumn{1}{c|}{{\color[HTML]{333333} 98.5\%}} & \cellcolor[HTML]{FFCCC9}{\color[HTML]{333333} 76.6\%} & \cellcolor[HTML]{FFCCC9}{\color[HTML]{333333} 70.5\%} & \cellcolor[HTML]{FFCCC9}{\color[HTML]{333333} 66.7\%} & {\color[HTML]{333333} 18.8} \\ \bottomrule
\end{tabular}%
}
\label{tab:spatial}
\end{table*}
\vspace{0.2em}
\textbf{Implementation details.}
We set the input size of images to be 256 x 256 that is the common output resolution of the existing synthetic model. 
For StatAttack, we set the parameters $\lambda_{a} = 0.1$, $\lambda_{\varphi} = 0.1$ and the iteration number is 40. For MStatAttack, we use 3 layers, and set the batch size to 30 for this experiment.

\subsection{Attack on Spatial-based Detectors} 
We conducte experiments to evaluate the effectiveness of our proposed attack on four space-based Deepfake detectors in both white-box and black-box settings.

\begin{figure}[t]
    \centering
    \includegraphics[width=\linewidth]{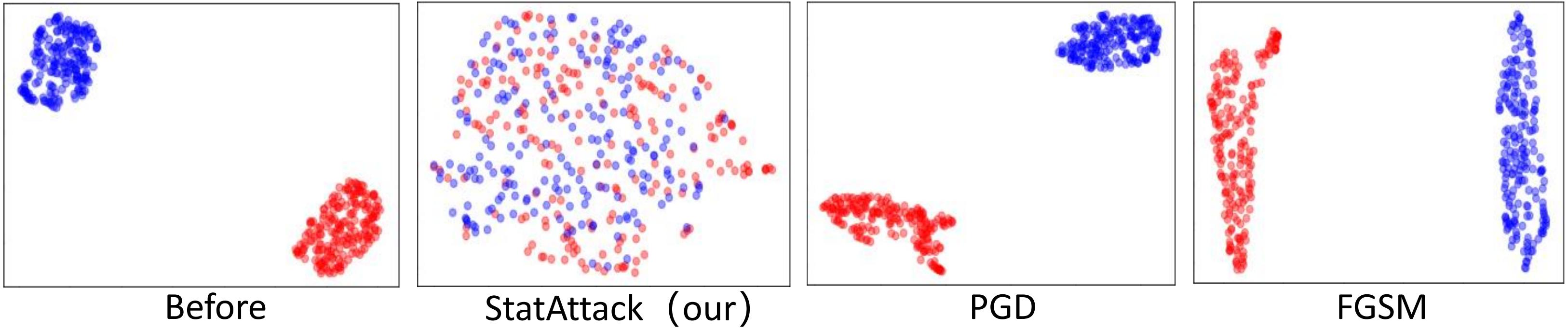}
    \caption{Feature distributions on ResNet after various attacks. Red and blue points represent feature distributions of attacked fake and real images, respectively. It is evident that the baseline attacks fail to align the feature distributions of fake and real images, whereas the proposed method can.}
    \label{fig:distribution}
\end{figure}

In the white-box setting, our attack achieves competitive results with baselines, as shown in Table \ref{tab:spatial}. Moreover, in comparison to StatAttack, MStatAttack is able to select a more efficient combination of perturbations, resulting in an improved attack success rate. For instance, when attacking ResNet on the StarGAN dataset, MStatAttack increases the attack success rate from 99.3$\%$ to 100$\%$. Additionally, we visualize the feature distributions of natural and attacked fake images using various attacks, as depicted in Figure \ref{fig:distribution}. It can be seen that our proposed method can align the feature distributions of fake and real images, making it more difficult for DeepFake detectors to classify them accurately.

In the black-box setting, we use adversarial examples generated by a specific detector to attack other detectors, enabling us to evaluate the transferability of our attack methods. Table \ref{tab:spatial} presents the results of our attacks that achieve higher transfer attack success rates than those of other methods with nearly top-2 results in all cases. For instance, when using adversarial examples crafted from ResNet on the DeepFake dataset, we achieve transfer attack success rates of 65.9$\%$, 81.3$\%$, and 71.2$\%$ on EfficientNet, DenseNet, and MobileNet, respectively.

\begin{table}[ht]
\centering
\caption{The attack success rate and image quality assessment on frequency-based Deepfake detectors. In each group, the first column denotes the attack success rates in the white-box setting, and the second columns are the attack success rates in the black-box setting. The ResNet column represents the transfer success rate of the adversarial examples generated by the ResNet. We mark the first, second, and third highest attack success rates in red, yellow, and blue. The last column shows the BRISQUE score. }
\resizebox{\linewidth}{!}{%
\begin{tabular}{@{}cc|ccc|ccc|ccc@{}}
\toprule
 & \cellcolor[HTML]{EFEFEF}Crafted from & \multicolumn{3}{c|}{\cellcolor[HTML]{EFEFEF}DCTA} & \multicolumn{3}{c|}{\cellcolor[HTML]{EFEFEF}DFTD} & \multicolumn{3}{c}{\cellcolor[HTML]{EFEFEF}ResNet} \\ \midrule
\multicolumn{1}{c|}{} & \cellcolor[HTML]{EFEFEF}Model\&ASR & \multicolumn{1}{c|}{\cellcolor[HTML]{EFEFEF}DCTA} & \cellcolor[HTML]{EFEFEF}DFTD & \cellcolor[HTML]{EFEFEF}BRIS & \multicolumn{1}{c|}{\cellcolor[HTML]{EFEFEF}DFTD} & \cellcolor[HTML]{EFEFEF}DCTA & \cellcolor[HTML]{EFEFEF}BRIS & \cellcolor[HTML]{EFEFEF}DCTA & \cellcolor[HTML]{EFEFEF}DFTD & \multicolumn{1}{l}{\cellcolor[HTML]{EFEFEF}BRIS} \\ \midrule
\multicolumn{1}{c|}{} & PGD & \multicolumn{1}{c|}{65.2\%} & 2.0\% & 61.2 & \multicolumn{1}{c|}{81.2\%} & 3.5\% & 59.6 & 0.1\% & {\color[HTML]{000000} 1.1\%} & 58.2 \\
\multicolumn{1}{c|}{} & FGSM & \multicolumn{1}{c|}{54.1\%} & 7.3\% & 60.1 & \multicolumn{1}{c|}{75.4\%} & 4.1\% & 63.4 & \cellcolor[HTML]{96FFFB}1.2\% & \cellcolor[HTML]{96FFFB}2.1\% & 62.3 \\
\multicolumn{1}{c|}{} & MIFGSM & \multicolumn{1}{c|}{72.3\%} & 7.5\% & 55.2 & \multicolumn{1}{c|}{83.7\%} & \cellcolor[HTML]{96FFFB}8.1\% & 57.1 & 0.8\% & 1.0\% & 57.7 \\
\multicolumn{1}{c|}{} & VMIFGSM & \multicolumn{1}{c|}{73.4\%} & \cellcolor[HTML]{96FFFB}8.6\% & 54.0 & \multicolumn{1}{c|}{85.6\%} & 6.3\% & 55.3 & 0.0\% & 0.0\% & 56.3 \\ \cmidrule(l){2-11} 
\multicolumn{1}{c|}{} & \textbf{StatAttack} & \multicolumn{1}{c|}{85.0\%} & \cellcolor[HTML]{FFFFC7}35.5\% & 58.8 & \multicolumn{1}{c|}{91.1\%} & \cellcolor[HTML]{FFCCC9}38.8\% & 57.6 & \cellcolor[HTML]{FFCCC9}18.5\% & \cellcolor[HTML]{FFCCC9}23.7\% & 59.0 \\
\multicolumn{1}{c|}{\multirow{-6}{*}{\rotatebox{90}{FF++}}} & \textbf{MStatAttack} & \multicolumn{1}{c|}{87.5\%} & \cellcolor[HTML]{FFCCC9}37.2\% & 57.1 & \multicolumn{1}{c|}{93.2\%} & \cellcolor[HTML]{FFFC9E}37.0\% & 57.3 & \cellcolor[HTML]{FFFFC7}17.3\% & \cellcolor[HTML]{FFFFC7}22.6\% & 53.8 \\ \midrule
\multicolumn{1}{c|}{} & PGD & \multicolumn{1}{c|}{71.2\%} & 0.0\% & 27.6 & \multicolumn{1}{c|}{88.8\%} & 3.5\% & 28.5 & 1.5\% & 1.3\% & 23.5 \\
\multicolumn{1}{c|}{} & FGSM & \multicolumn{1}{c|}{60.1\%} & \cellcolor[HTML]{96FFFB}1.2\% & 32.0 & \multicolumn{1}{c|}{54.4\%} & 7.2\% & 32.4 & \cellcolor[HTML]{96FFFB}3.3\% & \cellcolor[HTML]{96FFFB}4.5\% & 26.4 \\
\multicolumn{1}{c|}{} & MIFGSM & \multicolumn{1}{c|}{88.3\%} & 0.0\% & 26.2 & \multicolumn{1}{c|}{75.3\%} & 5.1\% & 26.3 & 0.7\% & 1.2\% & 24.8 \\
\multicolumn{1}{c|}{} & VMIFGSM & \multicolumn{1}{c|}{89.5\%} & 1.1\% & 31.1 & \multicolumn{1}{c|}{80.4\%} & \cellcolor[HTML]{96FFFB}7.5\% & 27.1 & 1.6\% & 2.1\% & 27.3 \\ \cmidrule(l){2-11} 
\multicolumn{1}{c|}{} & \textbf{StatAttack} & \multicolumn{1}{c|}{92.1\%} & \cellcolor[HTML]{FFFFC7}26.2\% & 28.2 & \multicolumn{1}{c|}{89.8\%} & \cellcolor[HTML]{FFFFC7}28.2\% & 27.0 & \cellcolor[HTML]{FFCCC9}17.3\% & \cellcolor[HTML]{FFFFC7}21.6\% & 27.4 \\
\multicolumn{1}{c|}{\multirow{-6}{*}{\rotatebox{90}{StyleGANv2}}} & \textbf{MStatAttack} & \multicolumn{1}{c|}{93.4\%} & \cellcolor[HTML]{FFCCC9}31.3\% & 26.4 & \multicolumn{1}{c|}{91.2\%} & \cellcolor[HTML]{FFCCC9}31.6\% & 26.7 & \cellcolor[HTML]{FFFFC7}12.1\% & \cellcolor[HTML]{FFCCC9}22.7\% & 23.0 \\ \midrule
\multicolumn{1}{c|}{} & PGD & \multicolumn{1}{c|}{71.2\%} & 2.2\% & 40.4 & \multicolumn{1}{c|}{91.0\%} & 3.2\% & 43.1 & 2.3\% & 3.6\% & 42.8 \\
\multicolumn{1}{c|}{} & FGSM & \multicolumn{1}{c|}{65.3\%} & 3.2\% & 41.4 & \multicolumn{1}{c|}{75.4\%} & 5.1\% & 51.2 & \cellcolor[HTML]{96FFFB}4.2\% & \cellcolor[HTML]{96FFFB}5.3\% & 51.1 \\
\multicolumn{1}{c|}{} & MIFGSM & \multicolumn{1}{c|}{87.3\%} & 8.0\% & 40.5 & \multicolumn{1}{c|}{93.2\%} & 8.2\% & 54.6 & 1.6\% & 2.1\% & 50.3 \\
\multicolumn{1}{c|}{} & VMIFGSM & \multicolumn{1}{c|}{86.5\%} & \cellcolor[HTML]{96FFFB}9.1\% & 43.9 & \multicolumn{1}{c|}{95.1\%} & \cellcolor[HTML]{96FFFB}9.3\% & 52.1 & 1.8\% & 2.0\% & 42.2 \\ \cmidrule(l){2-11} 
\multicolumn{1}{c|}{} & \textbf{StatAttack} & \multicolumn{1}{c|}{95.1\%} & \cellcolor[HTML]{FFFFC7}26.9\% & 42.4 & \multicolumn{1}{c|}{91.7\%} & \cellcolor[HTML]{FFFC9E}31.1\% & 55.8 & \cellcolor[HTML]{FFCCC9}19.5\% & \cellcolor[HTML]{FFFFC7}25.6\% & 41.5 \\
\multicolumn{1}{c|}{\multirow{-6}{*}{\rotatebox{90}{StarGAN}}} & \textbf{MStatAttack} & \multicolumn{1}{c|}{96.4\%} & \cellcolor[HTML]{FFCCC9}28.2\% & 41.2 & \multicolumn{1}{c|}{88.7\%} & \cellcolor[HTML]{FFCCC9}33.8\% & 45.1 & \cellcolor[HTML]{FFFFC7}15.7\% & \cellcolor[HTML]{FFCCC9}19.3\% & 39.0 \\ \midrule
\multicolumn{1}{c|}{} & PGD & \multicolumn{1}{c|}{52.1\%} & 2.0\% & 27.7 & \multicolumn{1}{c|}{93.3\%} & 2.1\% & 29.8 & 4.5\% & 7.9\% & 28.9 \\
\multicolumn{1}{c|}{} & FGSM & \multicolumn{1}{c|}{43.6\%} & \cellcolor[HTML]{96FFFB}3.5\% & 32.4 & \multicolumn{1}{c|}{81.4\%} & 1.2\% & 33.6 & \cellcolor[HTML]{96FFFB}7.1\% & \cellcolor[HTML]{96FFFB}8.3\% & 21.2 \\
\multicolumn{1}{c|}{} & MIFGSM & \multicolumn{1}{c|}{71.5\%} & 2.1\% & 27.2 & \multicolumn{1}{c|}{97.3\%} & \cellcolor[HTML]{96FFFB}8.6\% & 28.6 & 6.5\% & 7.6\% & 29.3 \\
\multicolumn{1}{c|}{} & VMIFGSM & \multicolumn{1}{c|}{72.3\%} & 1.7\% & 27.4 & \multicolumn{1}{c|}{98.5\%} & 5.7\% & 28.8 & \cellcolor[HTML]{96FFFB}7.1\% & 7.9\% & 27.1 \\ \cmidrule(l){2-11} 
\multicolumn{1}{c|}{} & \textbf{StatAttack} & \multicolumn{1}{c|}{88.5\%} & \cellcolor[HTML]{FFFFC7}18.0\% & 30.9 & \multicolumn{1}{c|}{97.5\%} & \cellcolor[HTML]{FFCCC9}21.0\% & 29.3 & \cellcolor[HTML]{FFCCC9}11.2\% & \cellcolor[HTML]{FFCCC9}18.5\% & 26.8 \\
\multicolumn{1}{c|}{\multirow{-6}{*}{\rotatebox{90}{ProGAN}}} & \textbf{MStatAttack} & \multicolumn{1}{c|}{90.2\%} & \cellcolor[HTML]{FFCCC9}20.3\% & 31.2 & \multicolumn{1}{c|}{97.7\%} & \cellcolor[HTML]{FFFFC7}18.1\% & 29.0 & \cellcolor[HTML]{FFFC9E}12.2\% & \cellcolor[HTML]{FFFC9E}16.6\% & 24.1 \\ \bottomrule
\end{tabular}%
}\label{tab:frenquency}
\end{table}

\subsection{Attack on Frequency-based Detectors}
To further demonstrate the attack capability of our method, we evaluate its effectiveness on two frequency-based detectors, namely DCTA \cite{frank2020leveraging} and DFTD \cite{zhang2019detecting}. As shown in the first column of Table \ref{tab:frenquency}, the attack success rates of our proposed attack are significantly higher than those of baseline attacks in both white-box and black-box settings.
Moreover, our method can alter the frequency components of the fake images, resulting in generated adversarial examples closer to natural images in terms of frequency information. 
Therefore, the adversarial examples crafted from the spatial-based detector can also transfer to the frequency-based detectors, leading to the transferability in a broader sense. 
To demonstrate this, we used adversarial examples crafted from a spatial-based detector to attack frequency-based detectors. 
As shown in the last three columns of Table \ref{tab:frenquency}, the transfer attack success rate of our proposed attack significantly outperforms that of the baseline attacks.

\begin{figure}[t]
    \centering
    \includegraphics[width=\linewidth]{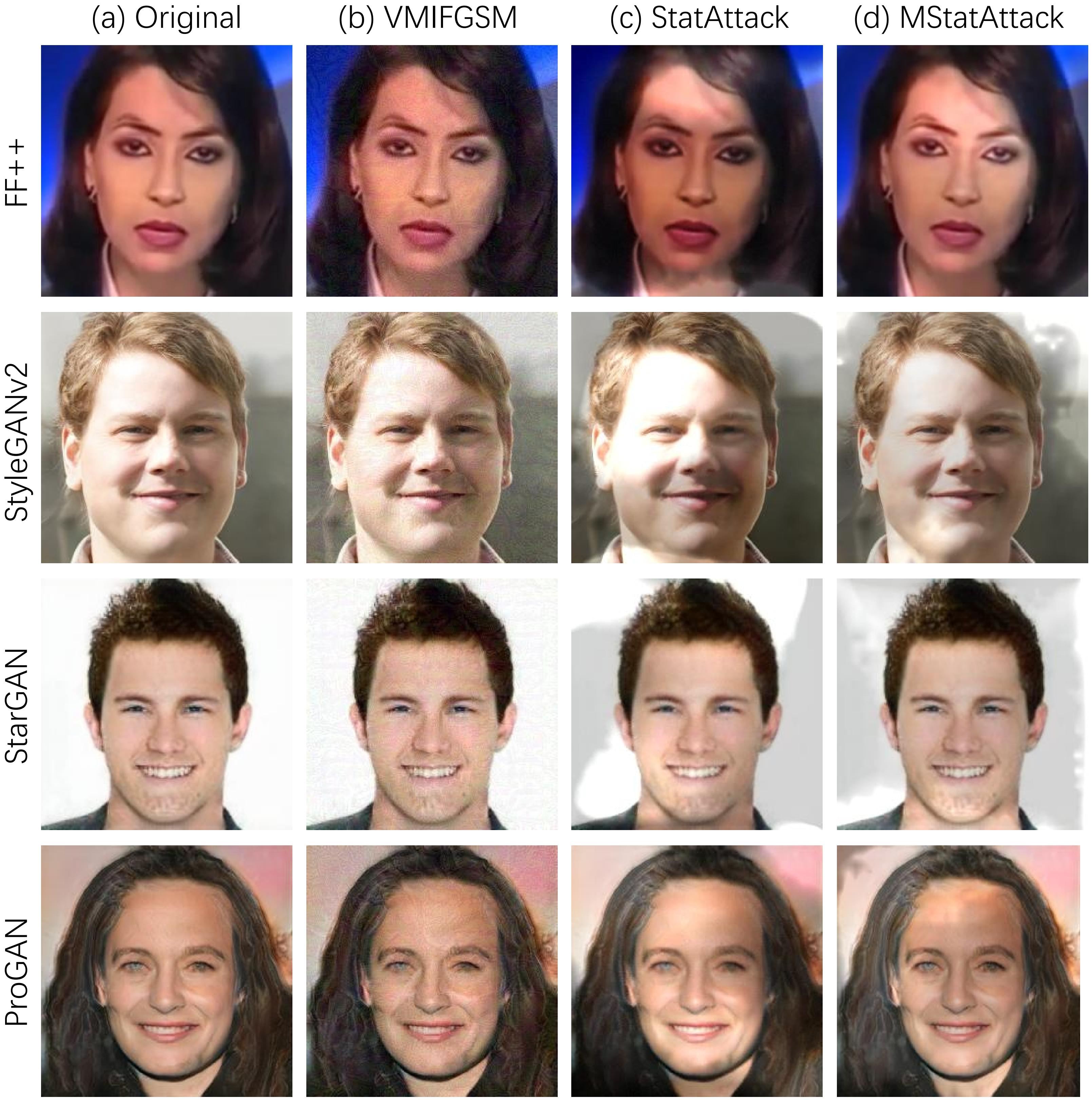}
    \caption{Visualization of the crafted adversarial examples. The visualizations include the raw fake images and the adversarial examples generated with VMIFGSM (b), StatAttack (c), and MStatAttack (d) on four datasets.}
    \label{fig:visualization}
\end{figure}
\subsection{Images Quality Comparison}
We also assess the image quality of the generated adversarial examples. The BRISQUE scores, shown in the last column of Tables \ref{tab:spatial} and \ref{tab:frenquency}, demonstrate that our attack method can maintain image quality while achieving high attack success rates. As illustrated in Figure \ref{fig:visualization}, our method (i.e., StatAttack) can generates natural-looking adversarial examples, and that MStatAttack can further enhance the realism of the adversarial examples compared to StatAttack.

\begin{table}[t]
\centering
\caption{Transfer success rate with different attack patterns. The adversarial examples are crafted from the ResNet.}
\resizebox{\linewidth}{!}{%
\begin{tabular}{@{}l|lll|ll@{}}
\toprule
\begin{tabular}[c]{@{}l@{}}Detector\&\\ Attack patterns\end{tabular} & EfficientNet & DenseNet & MobileNet & DCTA & DFTD \\ \midrule
only noise & 22.5\% & 38.6\% & 32.4\% & 3.2\% & 4.8\% \\
only exposure & 26.8\% & 35.3\% & 36.6\% & 11.4\% & 14.2\% \\
only blur & 10.3\% & 11.7\% & 16.5\% & 8.5\% & 10.3\% \\
w/o noise & 37.5\% & 31.0\% & 43.4\% & 17.1\% & 20.5\% \\
w/o exposure & 39.5\% & 33.7\% & 40.5\% & 10.5\% & 13.3\% \\
w/o blur & 57.1\% & 70.3\% & 65.4\% & 10.7\% & 12.9\% \\ \bottomrule
\end{tabular}%
}\label{tab:ablation}
\end{table}
\subsection{Ablation Study}
To assess the impact of each attack pattern on transferability, we perform transfer attack experiments using adversarial examples crafted from a spatial-based Deepfake detector ( \ie, ResNet ) in different attack patterns. Our results, as presented in Table \ref{tab:ablation}, reveal that adversarial noise and adversarial exposure have the greatest impact on the spatial-based detectors. On the other hand, for frequency domain-based detectors, adversarial exposure and adversarial blur are the primary factors influencing the attack performance.
\section{Conclusion}

In this work, we propose a novel degradation-based attack method called StatAttack, which can be used to bypass various DeepFake detectors. StatAttack is capable of generating adversarial examples that closely resemble natural images in terms of feature distributions.
In addition, we further extend StatAttack to a more powerful version, MStatAttack, which can  select a more effective combination for attack and generate more natural-looking adversarial examples.
We extensively evaluate our attack method on four spatial-based detectors and two frequency-based detectors using four datasets.
The experimental results demonstrate the effectiveness of our attack method in both white-box and black-box settings. 

{\bf Limitations.}
One limitation of our study is that we only considered three types of degradations. To enhance the robustness of our method, future work will focus on exploring additional natural degradations that could be more effective in simulating real-world scenarios. We also plan to leverage meta-learning to enhance the efficiency of our attack.

{\bf Social impacts.}
Our proposed attack method is capable of generating natural-looking adversarial examples that can transfer across different DeepFake detectors, posing a significant practical threat to the current DNN-based DeepFake detectors. This stealthy and transferable attack method can be employed to evaluate the robustness of Deepfake detectors in real-world applications or improve their robustness through adversarial training.
\section*{Acknowledgements}
This work was supported in part by JST SPRING Grant No.JPMJSP2136, JSPS KAKENHI Grant No.JP20H04168, No.JP21H04877, JST-Mirai Program Grant No.JPMJMI20B8, A*STAR Centre for Frontier AI Research, as well as Canada CIFAR AI Chairs Program, the Natural Sciences and Engineering Research Council of Canada.

{\small
\bibliographystyle{ieee_fullname}
\bibliography{egbib}
}

\newpage

\section{Supplementary Material}

In the main paper, we have reported the attack results on various DeepFake detectors and compared the transferability and image quality with four baseline attacks. 
In the supplementary material, (1) We describe in detail the polynomial model and the smoothing function used to generate the adversarial exposure.
(2) For the adversarial blur, we visualize the attack effect of adversarial Gaussian blur with different kernel sizes.
(3) We show more image quality comparisons with baseline attacks.  
(4) We provide more implementation details.

\subsection{Polynomial Model}
\label{sec:model}
The model provided by Gao \etal \cite{gao2022can} consists of two parts, the first part is a polynomial model for generating the exposure, and the second part is a smoothing formula that maintains the naturalness of the exposure. The polynomial model is shown below:
\begin{equation}
	\tilde{\mathbf{E_{i}}}=\sum_{t=0}^{D}\sum_{l=0}^{D-t}a_{t,l}T_{\varphi}\left(x_{i}\right )^{t}T_{\varphi}\left(y_{i}\right)^{l}
\end{equation}
Where the "$\tilde{\cdot}$" presents the logarithmic operations, $\left \{ a_{t,l} \right \}$ and $D$ denote the parameters and degree of the polynomial model, respectively. The number of parameters are $\left | \left \{ a_{t,l} \right \}  \right | =\frac{(D+1)(D+2)}{2}$, the lower degree $D$ leads to less model parameters $\left \{ a_{t,l} \right \}$. $T_{\varphi}\left(\cdot\right)$ is the offset transformation with $\varphi$ being the control points. We denote $i$ as the $i$-th pixel, and its corresponding coordinate as $\left ( x_{i},y_{i}  \right )$. The $\left ( T_{\varphi } \left ( x_{i} \right ), T_{\varphi } \left ( y_{i} \right )\right )$ indicates that the exposure field is warped by offsetting the control points of each coordinate. In addition, The polynomial model with fewer parameters $\left \{ a_{t,l} \right \}$ leads to a smoother exposure field. For convenient representations, we concatenate $\left \{ a_{t,l} \right \}$ as $\textbf{a}$. We can maintain the smoothness of the generated exposure by restricting the $\textbf{a}$  and $\varphi$, and the smoothing function can be written as follow:
\begin{equation}
	S\left ( \textbf{a},\varphi\right ) = -\lambda_{a}\left \| \textbf{a}\right \|_{2}^{2}-\lambda_{\varphi} \left \| \bigtriangledown \varphi  \right \|_{2}^{2}  
\end{equation}
The first term is to maintain the sparsity of $\textbf{a}$ and thus ensure the smoothness of the generated adversarial exposure. The second term is to force the control point $\varphi$ to vary in a smaller range, encouraging a minor warping of the adversarial exposure field. The hyper-parameters $\lambda_{a}$ and $\lambda_{b}$ are used to regulate the balance between adversarial attack and smoothness. In detail, we initialize $\textbf{a}$ be a vector whose entries share a common value (e.g., 0), the $\textbf{a}$ and $\varphi$ are then gradually optimized in each attack iteration to produce smooth and effective adversarial exposure.

\subsection{Different Kernel Sizes}
\label{sec:Kernal Size}
\begin{figure}[ht]
	\centering
	\includegraphics[width=\linewidth]{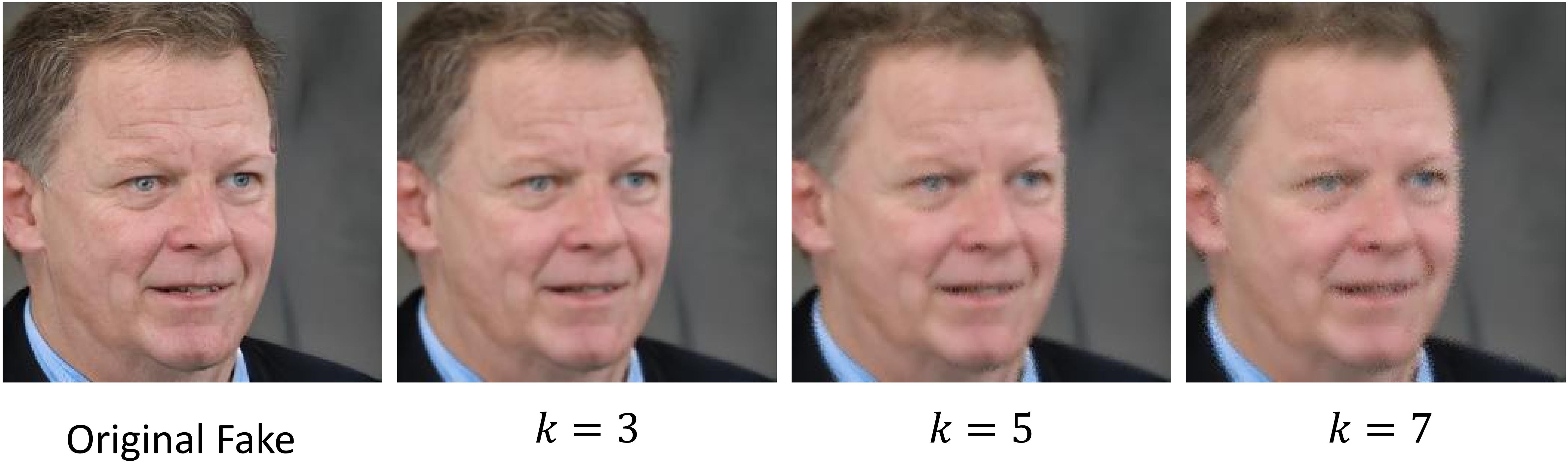}
	\caption{Adversarial examples generated by adversarial Gaussian blur with different kernel sizes. From left to right are the original image and the generated adversarial example with the Gaussian kernel size $k$ set to 3,5, and 7, respectively.}
	\label{fig:blurk}
\end{figure}

The success of adversarial Gaussian blur is dependent on the size of the Gaussian kernel and the value of $\sigma$. While a larger Gaussian kernel and $\sigma$ can improve the success rate of transfer attacks, they can also result in more blurry images. To illustrate this point, we present a visualization of the adversarial examples generated by adversarial Gaussian blur using different kernel sizes, as shown in Figure \ref{fig:blurk}.
\subsection{Additional Quality Comparisons}
\label{sec:quality}
\begin{figure*}[ht]
	\centering
	\includegraphics[width=\linewidth]{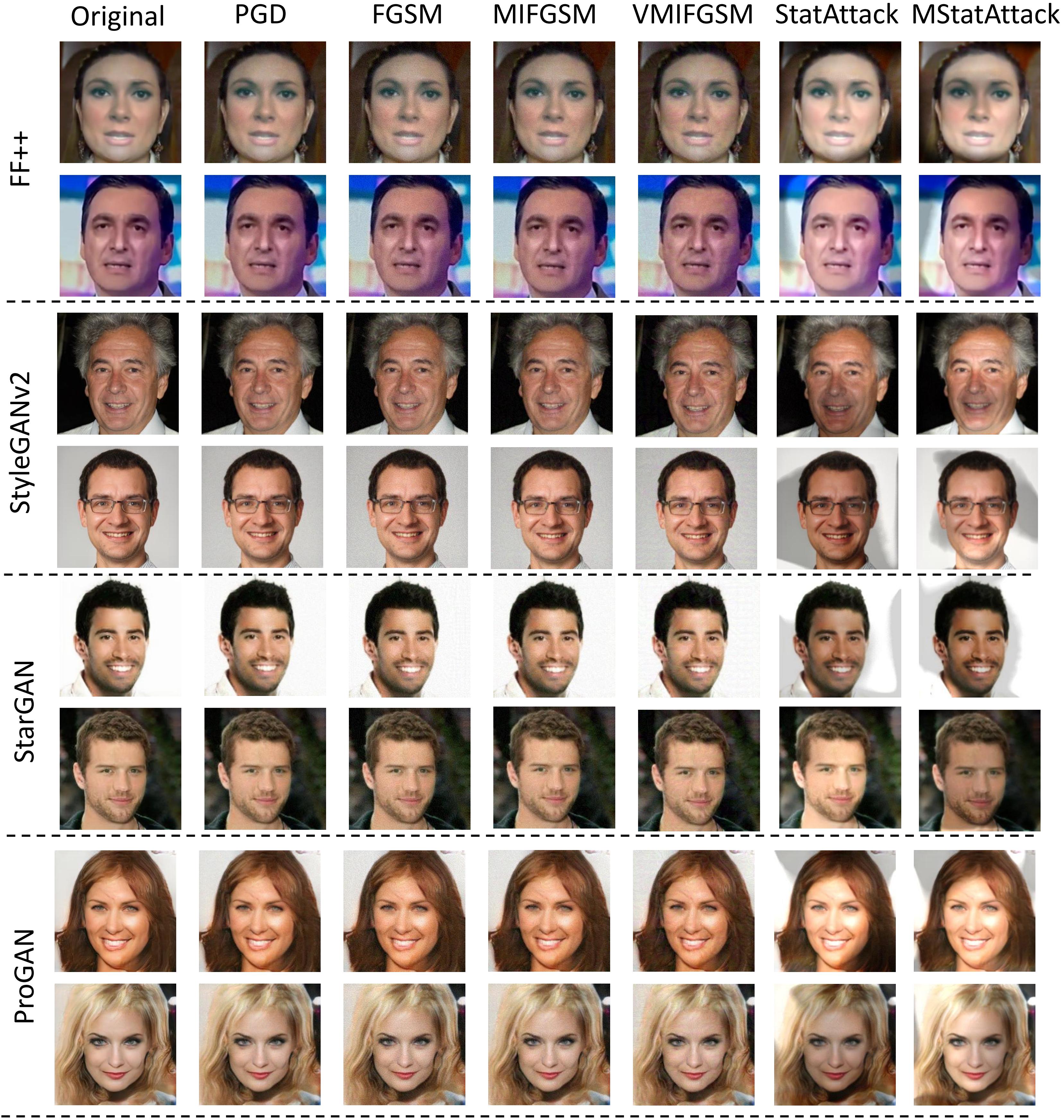}
	\caption{Visualization of generated adversarial examples. It contains the original fake image and the generated adversarial examples with the PGD, FGSM, MIFGSM, VMIFGSM, StatAttack, and MStatAttack on four datasets. }
	\label{fig:Mquality}
\end{figure*}

As shown in  Figure \ref{fig:Mquality}, we provide more quality comparisons of the adversarial samples generated with the baseline attacks \ie  PGD\cite{madry2017towards}, FGSM\cite{goodfellow2014explaining}, MIFGSM\cite{dong2018boosting} and VMIFGSM\cite{wang2021enhancing}.
\subsection{More Implementation Details}
\label{sec:implementataion}
In our experiments,  To balance the effectiveness of the attack and the quality of generated adversarial examples, we set the degree (\ie $D$) of the polynomial model to 11, the learning rate of $\textbf{a}$ and $\varphi$ is $10^{-2}$ and $10^{-3}$, respectively. For adversarial Gaussian blur, we set the Gaussian kernel size $k$ to 3. All the experiments were performed on a server running Ubuntu 6.0.8-arch1-1 system on a 10-core 3.60 GHz i9-10850k CPU with 31 GB RAM and an NVIDIA GeForce RTX 3090 GPU with 24 GB memory.

\end{document}